\documentclass[10pt,twocolumn,letterpaper]{article}

\usepackage{iccv}
\usepackage{times}
\usepackage{epsfig}
\usepackage{graphicx}
\usepackage{amsmath}
\usepackage{amssymb}
\usepackage{pifont}
\usepackage{xcolor}
\usepackage{pgfplots}
\usepackage{multirow}
\usepackage{tikz} 
\usepackage[colorinlistoftodos]{todonotes}
\usepackage{pifont}
\usepackage{tabularx}
\usepackage{tabu}
\usepackage{comment}
\usepackage{caption}
\usepackage{subcaption}
\usepackage{graphics}

\usepackage[pagebackref=true,breaklinks=true,letterpaper=true,colorlinks,bookmarks=false]{hyperref}

\iccvfinalcopy 

\def\iccvPaperID{3281} 
\def\Shopagon{Varied Clothing Dataset}
\newcommand*\rfrac[2]{{}^{#1}\!/_{#2}}

\ificcvfinal\fi
\begin{document}


\title{IMP: Instance Mask Projection for High Accuracy Semantic Segmentation of Things}

\author{Cheng-Yang Fu \quad Tamara L. Berg
\quad Alexander C. Berg\\
Facebook AI\\
}

\maketitle

\begin{abstract}
In this work, we present a new operator, called Instance Mask Projection (IMP), which projects a predicted Instance Segmentation as a new feature for semantic segmentation. It also supports back propagation so is trainable end-to-end. Our experiments show the effectiveness of IMP on both Clothing Parsing (with complex layering, large deformations, and non-convex objects), and on Street Scene Segmentation (with many overlapping instances and small objects). On the Varied Clothing Parsing dataset (VCP), we show instance mask projection can improve 3 points on mIOU from a state-of-the-art Panoptic FPN segmentation approach. On the  ModaNet clothing parsing dataset, we show a dramatic improvement of 20.4\% absolutely compared to existing baseline semantic segmentation results. In addition, the instance mask projection operator works well on other (non-clothing) datasets, providing an improvement of 3 points in mIOU on Thing classes of Cityscapes, a self-driving dataset, on top of a state-of-the-art approach.

\end{abstract}


\vspace{-0.2cm}
\section{Introduction}
\vspace{-0.1cm}

This paper addresses producing pixel-accurate semantic segmentations.  This is relevant for a wide range of applications, from self-driving, where predicting accurate localizations of objects, buildings, people, etc, (as illustrated in the Cityscapes dataset~\cite{Cordts2016Cityscapes}), will be necessary for producing safe autonomous vehicles, to commerce, where accurate segmentations of the clothing items someone is wearing~\cite{zheng2018modanet} will form a foundational building block for applications like visual search. Many other potential applications can be envisioned, especially in real-world scenarios where intelligent agents are using vision to perceive their surrounding environments, but for this paper we focus on two areas, street scenes and fashion outfits, as two widely differing settings to demonstrate the generality of our method. 
    
\begin{figure}[t]
    \centering
    \includegraphics[width=0.49\textwidth]{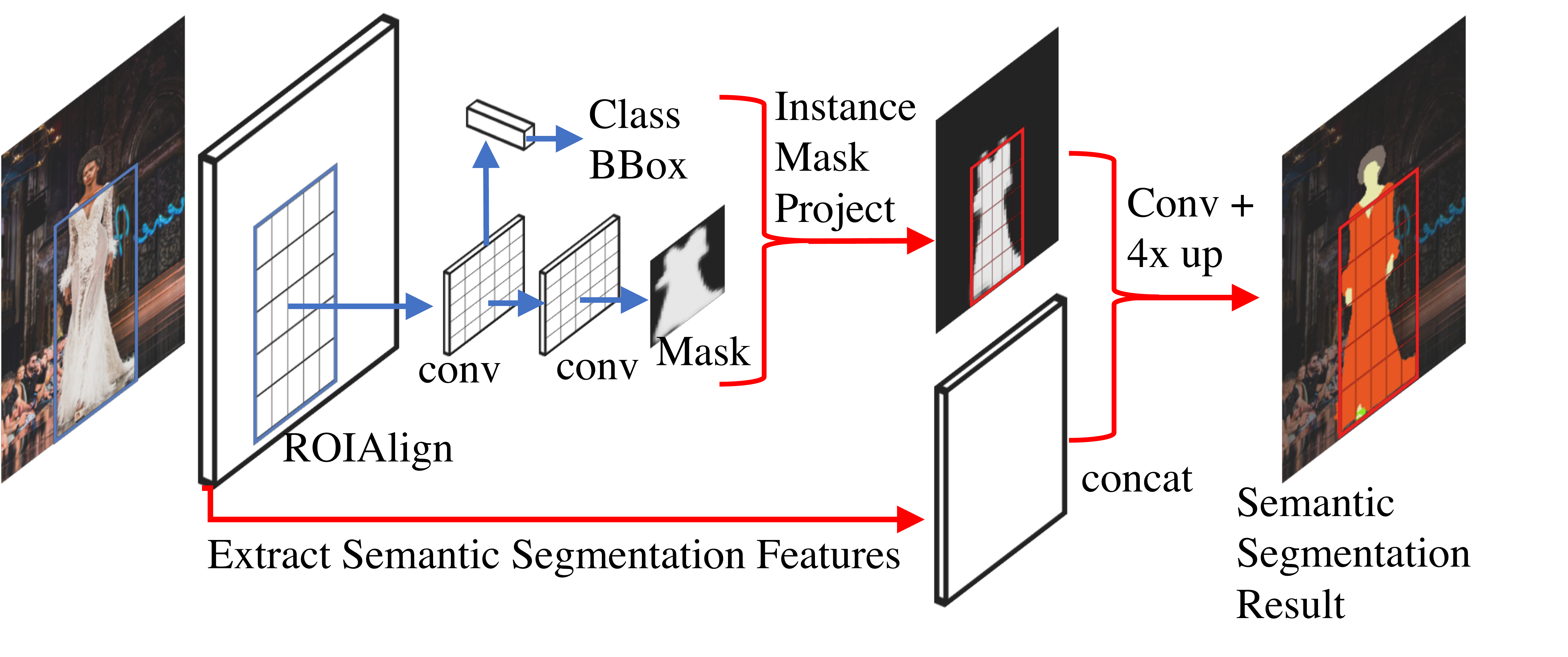}
    \caption{Example of Instance Mask Projection:  An Instance Mask Projection operator takes the instance mask as the input (Class, Score, BBox, Mask) and project the results as the feature map for semantic segmentation prediction.
    In this example, the ``Dress" is detected in the Instance Detection pipeline, then is transformed as the feature layer.}
\label{fig:imp_operator}
\end{figure}


We propose combining information from detection results, bounding box and instance mask prediction,as in Mask R-CNN~\cite{he2017maskrcnn}. The core of our approach is a new operator, Instance Mask Projection (IMP), that projects the predicted masks (with uncertainty) from Mask R-CNN for each detection into a feature map to use as an auxiliary input for semantic segmentation, significantly increasing accuracy.  Furthermore, in our implementations the semantic segmentation pipeline shares a trunk with the detector, as in Panoptic FPN~\cite{kirillov2019panopticfpn}, resulting in a fast solution.
    
This approach is most helpful for improving semantic segmentation of objects for which detection works well, movable foreground objects (things) as opposed to regions like grass (stuff).  Using the instance mask output from a detector allows the approach to make decisions about the presence/absence/category of an object as a unit, and to explicitly estimate and use the scale of a detected object for aggregating features (e.g. in roi-pooling).  In contrast, semantic segmentation must make the decision about object type over and over again at each location using a fixed scale for spatial context.  The semantic segmentation prediction deals better with concave shapes than the instance mask prediction, in addition to offering high-resolution output. 


As part of validating the effectiveness of this approach we demonstrate several new results:

    \begin{itemize}
        \vspace{-.1cm}
        \item The object masks predicted by Mask R-CNN~\cite{he2017maskrcnn} are sometimes more accurate than semantic segmentation for some objects. See Sec.~\ref{sec:experimentsVCD} and ~\ref{sec:experimentsModaNet}.
        \vspace{-.1cm}
        \item Following this insight we design the Instance Mask Projection (IMP) operator to project these masks as a feature for semantic segmentation, see Sec.~\ref{sec:imp}.
        \vspace{-.1cm}
        \item Segmentation results with IMP significantly improve on the state of the art for semantic segmentation on clothing segmentation.  Showing the best results on ModaNet~\cite{zheng2018modanet}, improving mean IOU from 51\% for DeepLabV3+ to 71.4\%. 
        See sec.~\ref{sec:experimentsModaNet}.
        \vspace{-.1cm}
        \item Across three datasets, using features from IMP improves significantly over a Panoptic segmentation baseline (the same system without IMP) and produces state of the art results.  See Sec.~\ref{sec:experimentsCityscapes}. 
        \vspace{-3.0mm}
    \end{itemize}
    
    


\begin{figure}[th!]

    \begin{subfigure}{0.49\textwidth}
        \centering
        \begin{subfigure}[b]{0.241\textwidth}
            \caption*{Input}
            \includegraphics[height=3cm]{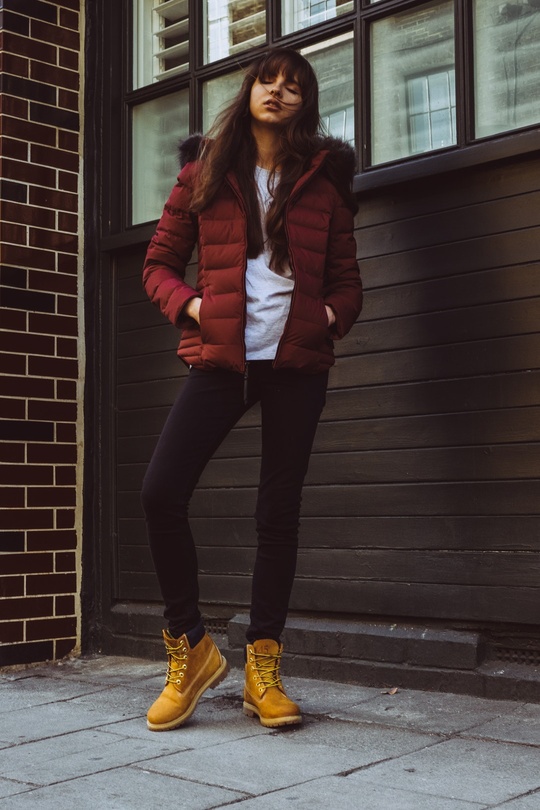}
        \end{subfigure}
        \hfill
        \begin{subfigure}[b]{0.241\textwidth}
            \caption*{\scriptsize Panoptic-FPN}
            \includegraphics[height=3cm]{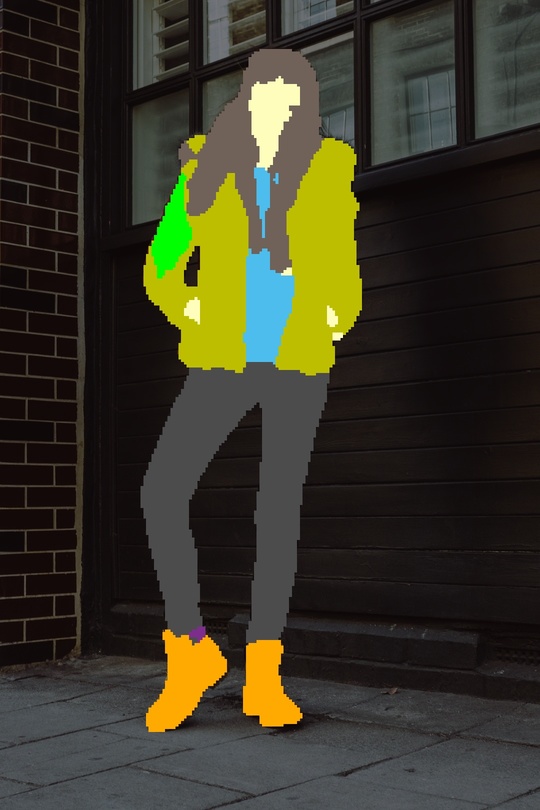}
        \end{subfigure}
        \hfill
        \begin{subfigure}[b]{0.241\textwidth}
            \caption*{\scriptsize Mask R-CNN-IMP}
            \includegraphics[height=3cm]{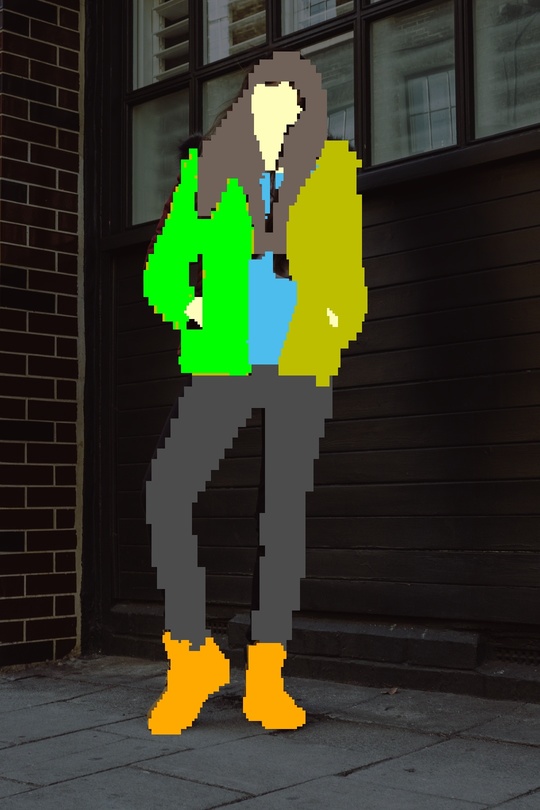}
        \end{subfigure}
        \hfill
        \begin{subfigure}[b]{0.241\textwidth}
            \caption*{\scriptsize Panoptic-FPN-IMP}
            \includegraphics[height=3cm]{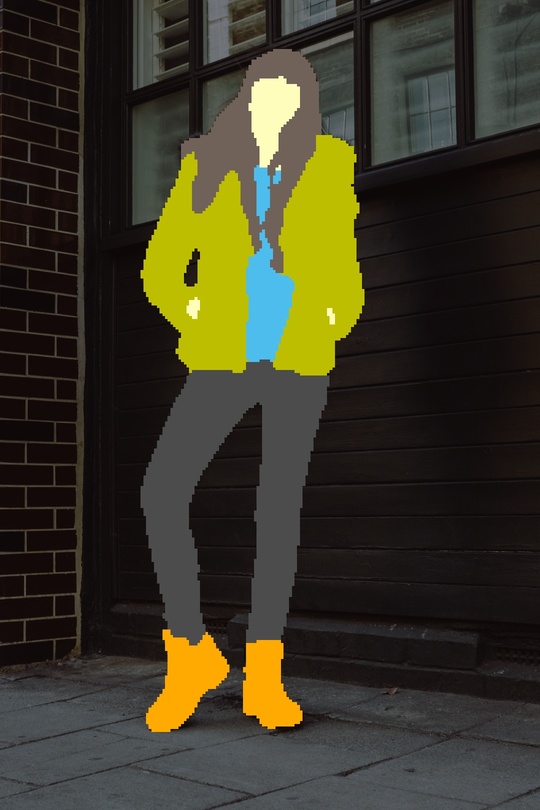}
        \end{subfigure}
    \caption{} 
    \label{fig:vis_girl_in_jacket}
    \end{subfigure}
    
    \begin{subfigure}{0.49\textwidth}
        \includegraphics[height=3cm]{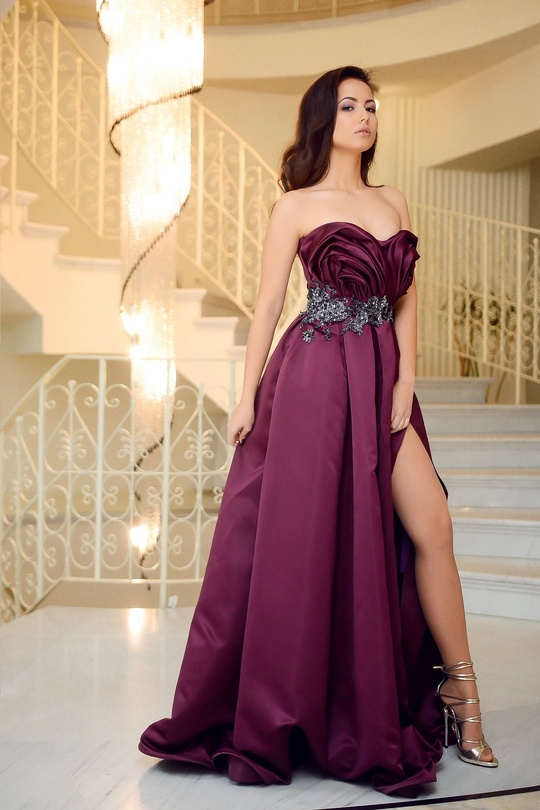} 
        \hfill
        \includegraphics[height=3cm]{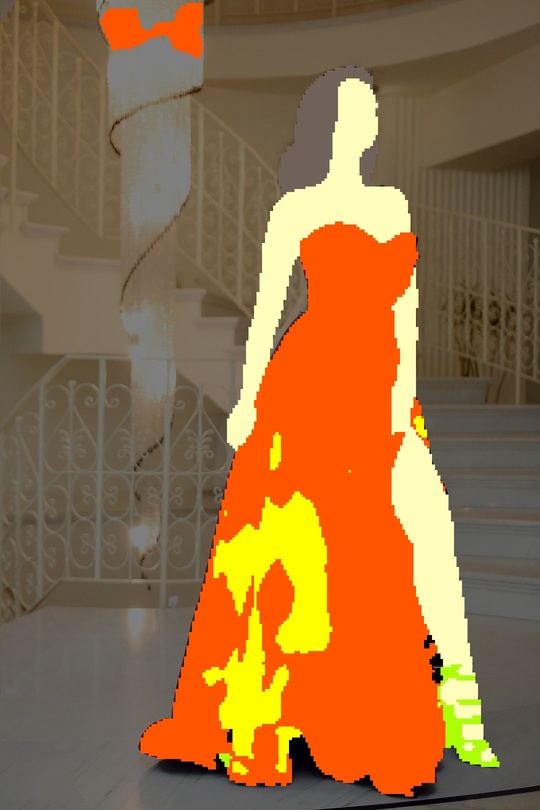}
        \hfill
        \includegraphics[height=3cm]{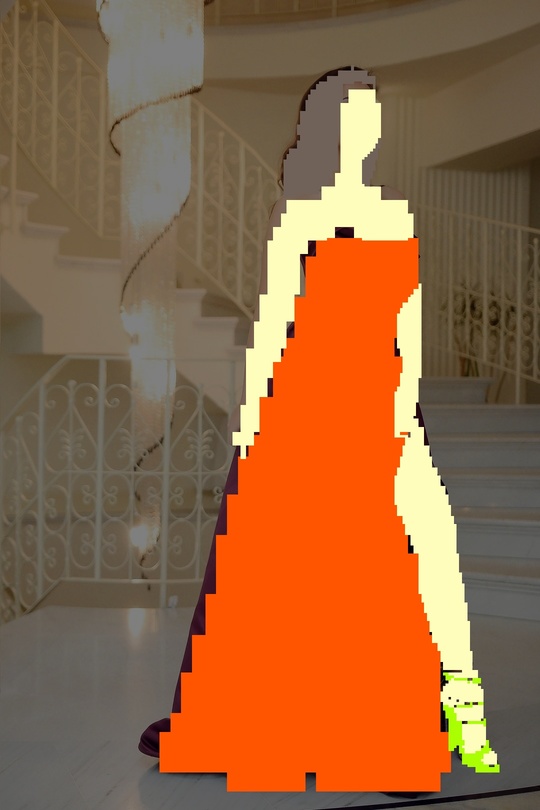}
        \hfill
        \includegraphics[height=3cm]{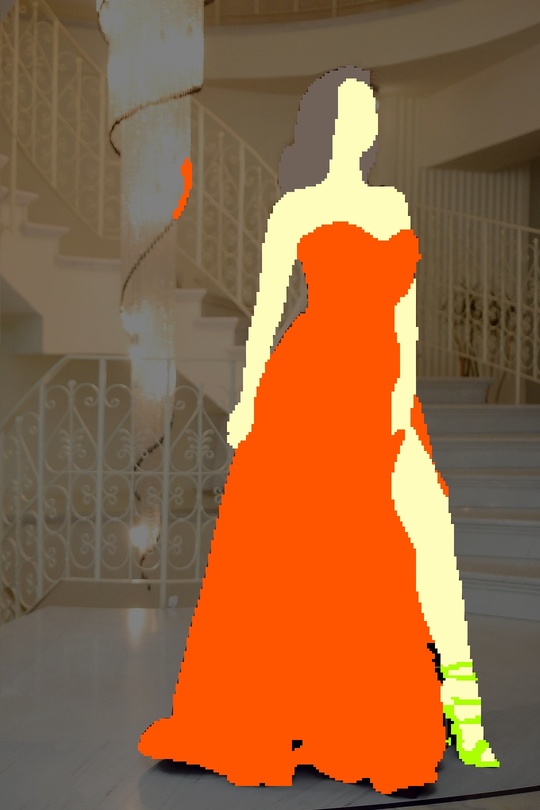}
    \caption{}
    \label{fig:vis_woman_in_dress}
    \end{subfigure}
    
    \begin{subfigure}{0.49\textwidth}
        \includegraphics[height=3cm]{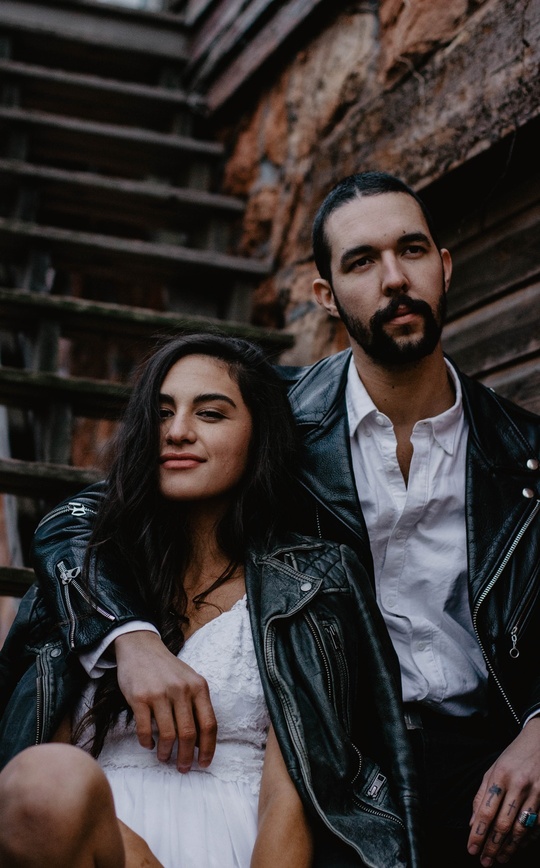}
        \hfill
        \includegraphics[height=3cm]{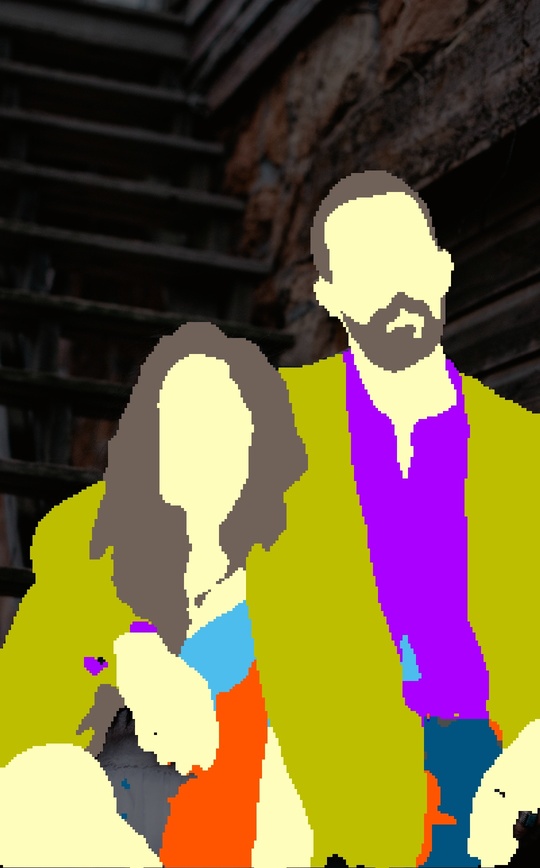}
        \hfill
        \includegraphics[height=3cm]{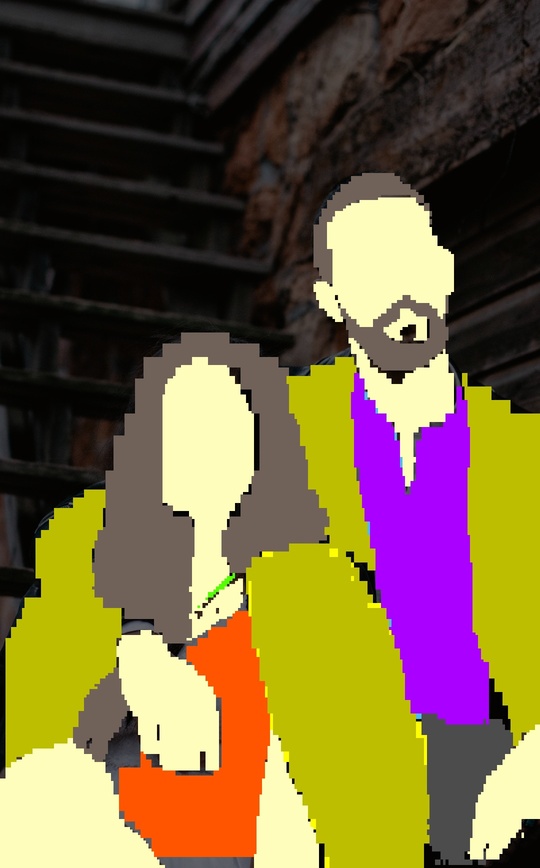}
        \hfill
        \includegraphics[height=3cm]{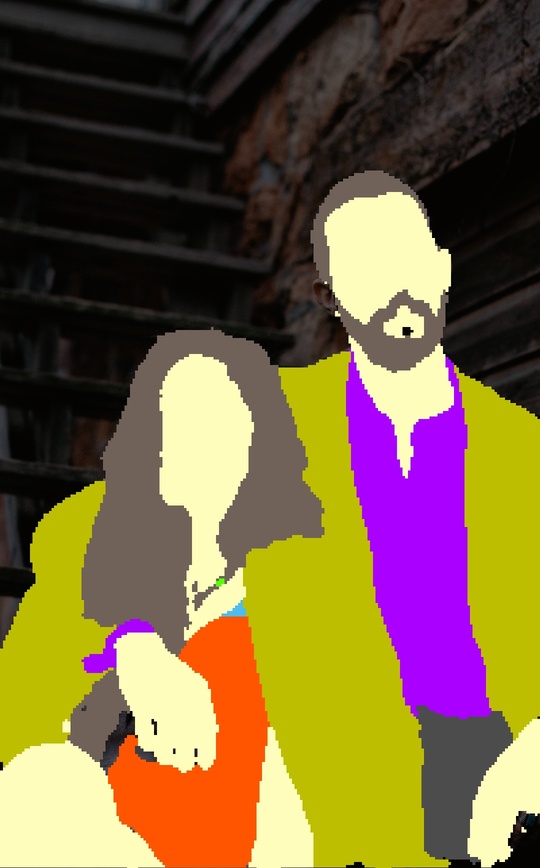}
    \caption{}
    \label{fig:vis_two}
    \end{subfigure}

    

    \begin{subfigure}{0.49\textwidth}
        \includegraphics[width=0.24\textwidth]{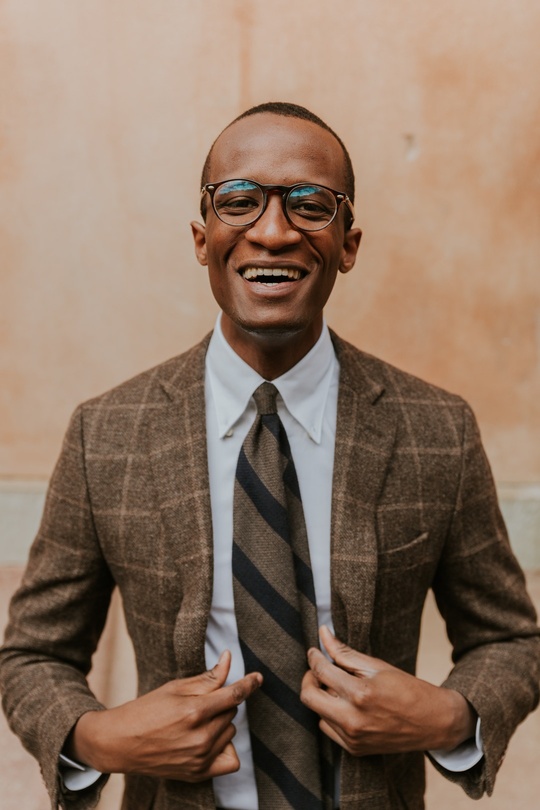}
        \includegraphics[width=0.24\textwidth]{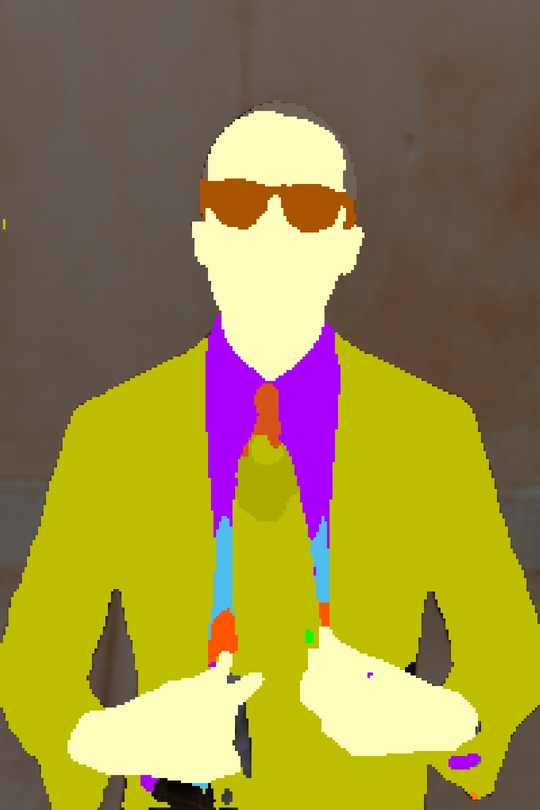}
        \includegraphics[width=0.24\textwidth]{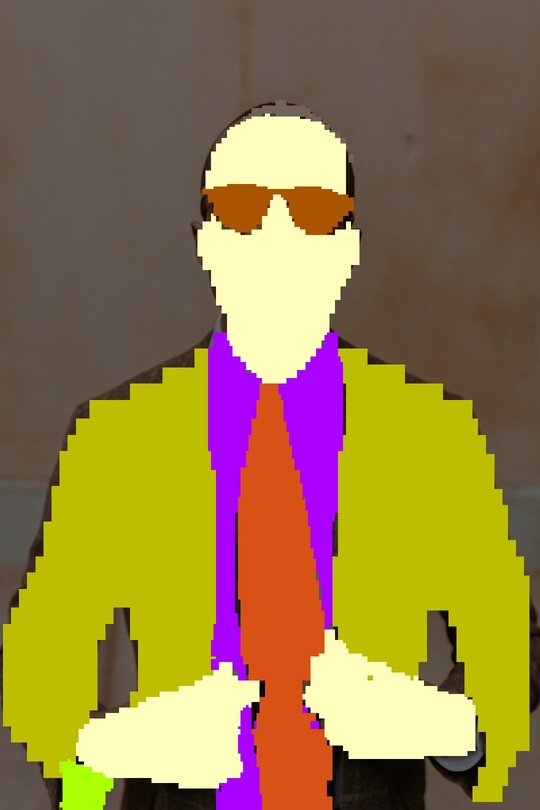}
        \includegraphics[width=0.24\textwidth]{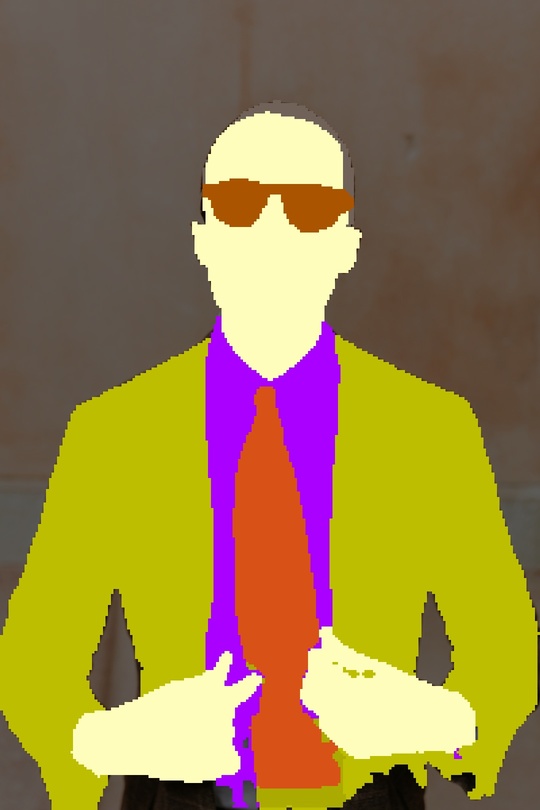}
    \caption{}
    \label{fig:vis_man_wears_tie}
    \end{subfigure}
    

    \begin{tikzpicture}[y=.5cm, x=.5cm]
	legend
	
	\draw[black!30] (0, -1) -- (17, -1);
	\draw[black!30] (0, 0) -- (17, 0);
	

    \draw[ fill={rgb,255:red,255; green,254; blue,189} ] (0,0) rectangle (1, 1);
    \draw (1, 0.5) node[right, font=\small]{skin};
    
    \draw[ fill={rgb,255:red,112; green,98; blue,90} ] (3,0) rectangle (4, 1);
    \draw (4, 0.5) node[right,font=\small]{hair};
    
    \draw[ fill={rgb,255:red,255; green,128; blue,0} ] (6,0) rectangle (7, 1);
    \draw (7, 0.5) node[right,font=\small]{hat};
    
    \draw[ fill={rgb,255:red,217; green,83; blue,25} ] (9,0) rectangle (10, 1);
    \draw (10, 0.5) node[right,font=\small]{tie};    
    
    \draw[ fill={rgb,255:red,170; green,85; blue,0} ] (12,0) rectangle (13, 1);
    \draw (13, 0.5) node[right,font=\footnotesize]{glasses};
    
    \draw[ fill={rgb,255:red,84; green,255; blue,0}  ] (15,0) rectangle (16, 1);
    \draw (16, 0.5) node[right,font=\footnotesize]{necklace};

    \draw[ fill={rgb,255:red,77; green,190; blue,238} ] (0,-1) rectangle (1, 0);
    \draw (1, -0.5) node[right,font=\small]{t-shirt};
    
    \draw[ fill={rgb,255:red,170; green,0; blue,255} ] (3,-1) rectangle (4, 0);
    \draw (4, -0.5) node[right,font=\small]{shirt};
    
    \draw[ fill={rgb,255:red,255; green,84; blue,0} ] (6,-1) rectangle (7, 0);
    \draw (7, -0.5) node[right,font=\small]{dress};
    
    \draw[ fill={rgb,255:red,191; green,191; blue,0} ] (9,-1) rectangle (10, 0);
    \draw (10, -0.5) node[right,font=\small]{jacket};
    
    \draw[ fill={rgb,255:red,0; green,255; blue,0} ] (12,-1) rectangle (13, 0);
    \draw (13, -0.5) node[right,font=\small]{coat};

    \draw[ fill={rgb,255:red,170; green,255; blue,0} ] (0,-2) rectangle (1, -1);
    \draw (1, -1.5) node[right,font=\small]{shoes};
    
    \draw[ fill={rgb,255:red,255; green,170; blue,0} ] (3, -2) rectangle (4, -1);
	\draw (4, -1.5) node[right,font=\small]{boots};
	
	\draw[ fill={rgb,255:red,77; green,77; blue,77}  ] (6,-2) rectangle (7, -1);
	\draw (7, -1.5) node[right,font=\small]{pants};
	
	\draw[ fill={rgb,255:red,0; green,0; blue,255} ] (9,-2) rectangle (10, -1);
    \draw (10, -1.5) node[right,font=\footnotesize]{leggings};
    
    \draw[ fill={rgb,255:red,255; green,255; blue,0} ] (12,-2) rectangle (13, -1);
    \draw (13, -1.5) node[right,font=\footnotesize]{jumpsuit};
    
\end{tikzpicture}
    \caption{From left to right, images, results of Panoptic-FPN, results of Mask R-CNN-IMP, results of our final model, Panoptic-FPN-IMP. Figure~\ref{fig:vis_woman_in_dress}, Figure~\ref{fig:vis_two} and ~\ref{fig:vis_man_wears_tie} show Mask R-CNN-IMP generates cleaner results than Panoptic-FPN. Figure~\ref{fig:vis_girl_in_jacket} shows combing semantic segmentation features and IMP can fix problems happened in both. Figure~\ref{fig:vis_woman_in_dress} shows Mask R-CNN-IMP causes less false positives. 
    The visualization images are not from either ~\Shopagon~ nor ModaNet~\cite{zheng2018modanet} to avoid potential copyright questions. All images shown are licensed. See more examples in Figure~\ref{fig:vis_cases_sup}.
    }
    \label{fig:vis_cases}
    \vspace{-3.0mm}
\end{figure}

\vspace{-0.1cm}
\section{Related Work}
\vspace{-0.1cm}
Our work builds on current state-of-the-art object detection and semantic segmentation models which have benefited greatly from recent advances in convolution neural network architectures. In this section, we first review recent progress on object localization and semantic segmentation. Then, we describe how our proposed model fits in with other works which integrate both object detection and semantic segmentation. 

\vspace{-0.1cm}
\subsection{Localizing Things}
\vspace{-0.1cm}
Initially, methods to localize objects in images mainly focused on predicting a tight bounding box around each object of interest. As the accuracy matured, research in object localization has expanded to not only produce a rectangular bounding box but also an instance segmentation, identifying which pixels corresponding to each object. 


\smallskip
\noindent
{\bf Object Detection:}
R-CNN~\cite{girshick2014rcnn} has been one of the most foundational lines of research driving recent developments in detection, initiating work on using the feature representations learned in CNNs for localization. Many related works continued this progress in two-stage detection approaches, including SPP Net~\cite{he2015sppnet}, Fast R-CNN, ~\cite{girshick15fastrcnn} and Faster R-CNN~\cite{ ren2015faster}. 
In addition, single-shot detectors YOLO~\cite{redmon2015yolo}, SSD~\cite{liu2016ssd} have been proposed to achieve real-time speed. Many other recent methods have been proposed to improve accuracy. R-FCN~\cite{dai2016rfcn} pools position-sensitive class maps to make predictions more robust. FPN~\cite{lin2016fpn} and DSSD~\cite{fu2017dssd} add top-down connections to bring semantic information from deep layers to shallow layers. FocalLoss~\cite{lin2017focal} reduces the extreme class imbalance by decreasing influence from well-predicted examples.  

\smallskip
\noindent
{\bf Instance Segmentation:}
Compared to early instance segmentation works~\cite{dai2016instance, li2016fully}, Mask R-CNN~\cite{he2017maskrcnn} identifies the core issue for mask prediction as ROI-pooling box misalignment and proposes a new solution, ROI-Alignment using bilinear interpolation to fix quantization error. Path Aggregation Network~\cite{liu2018path_aggregation_network} pools results on multiple layers rather than one to further improve results. 

\vspace{-0.2cm}
\subsection{Semantic Segmentation}
\vspace{-0.1cm}
Fully Convolutional Networks (FCN)~\cite{Shelhamer2016FCN} has been the foundation for many recent semantic segmentation models. FCN uses convolution layers to output semantic segmentation results directly. Most current semantic segmentation approaches can be roughly categorized into two types, dilated convolution, or encoder-decoder based methods. We describe each, and graphical model enhancements below.

\smallskip
\noindent
{\bf Dilated Convolution:}
Dilated convolution~\cite{Yu2016dilatedConv,chen2018deeplab} increases the dilated kernels to 
learn larger receptive fields with fewer convolutions, producing large benefits in semantic segmentation tasks where long range context is useful. Thus, many recent approaches~\cite{chen2018depthwise_atrous_conv,zhao2018psanet,zhao2017pspnet, rotabulo2018place} have incorporated dilated convolution. Deformable Convolution Network~\cite{dai17deformableconvolution} takes this idea one step further, 
learning to predict the sampling area to improve the convolution performance instead of using a fixed geometric structure.

\smallskip
\noindent
{\bf Encoder-Decoder Architecture:}
SegNet~\cite{Badrinarayanan2017segnet} and U-NET~\cite{Ronneberger2015unet} proposed adding a  decoder stage, to upsample the feature resolution and produce higher resolution semantic segmentations. Encoder-decoder frameworks have also been widely adopted in other localization related areas of computer vision, such as Facial Landmark Prediction~\cite{honari2016recombinator_networks}, Human Key Point Detection~\cite{Newell2016StackedHourGlasses}, Instance Segmentation~\cite{Pinheiro2016SharpMask}, and Object Detection~\cite{lin2016fpn,fu2017dssd} .

\smallskip
\noindent
{\bf Graphical Models:}
Although deep learning approaches have improved semantic segmentation results dramatically, the output result is often still not sharp enough. One common approach to alleviate these issues is to apply a CRF-based approach to make the output more aligned with the color differences. Fully connected CRF~\cite{chen2018depthwise_atrous_conv,chen2015deeplabcrf}, and Domain Transform~\cite{cheng2016domaintransform} are two such approaches that can be trained with  neural networks in an end-to-end manner.  Soft Segmentation~\cite{aksoy2018semanticsoftsegmentation} fuses high-level semantic information with low-level texture and color features to carefully construct a graph structure, whose   corresponding Laplacian matrix and its eigenvectors reveal the semantic objects and the soft transitions between them. Soft segments can then be generated via eigen decomposition.  
Although using graphical models can make the prediction boundary align with the color differences, it cal also cause small objects to disappear due to excessive smoothing. Additionally, these methods all rely on good semantic segmentation results.

\vspace{-0.1cm}
\subsection{Combined Detection \& Semantic Segmentation} 
\vspace{-0.1cm}
In part due to newly released datasets, such as COCO-Stuff~\cite{caesar2018cocostuff}, research efforts toward integrating object detection/instance segmentation and semantic segmentation in a single network have increased. Panoptic Segmentation~\cite{kirillov2017panoptic} proposed a single evaluation metric to integrate instance segmentation and semantic segmentation. Following these efforts, Panoptic FPN~\cite{kirillov2019panopticfpn} showed that the FPN architecture can easily integrate both tasks in one network trained end-to-end.  Earlier work, Blitznet~\cite{dvornik17blitznet}, also demonstrated that both tasks can be improved in multitask training. One related improvement on Panoptic FPN is UPSNet~\cite{xiong2019upsnet}.  This uses a projection like our instance mask projection for a different purpose.  UPSNet~\cite{xiong2019upsnet} uses projected instance masks stacked with semantic segmentation outputs to make decision about which type of prediction (an instance mask or semantic segmentation) to use at each location.  This decision is made using softmax (without learning).  Instead our approach uses the projected instance masks as features to improve semantic segmentation, as orthogonal improvement.

Although we use Mask R-CNN~\cite{he2017maskrcnn} / Panoptic FPN~\cite{kirillov2019panopticfpn} architectures for producing instance segmentation and semantic segmentation predictions, our mask project operator is general and could alternatively make use of other instance and semantic segmentation methods as baseline models. Our method can easily take advantage of future development on both tasks to provide better combined results. 

\begin{figure}[th]
    \begin{subfigure}{0.49\textwidth}
        \includegraphics[width=\textwidth]{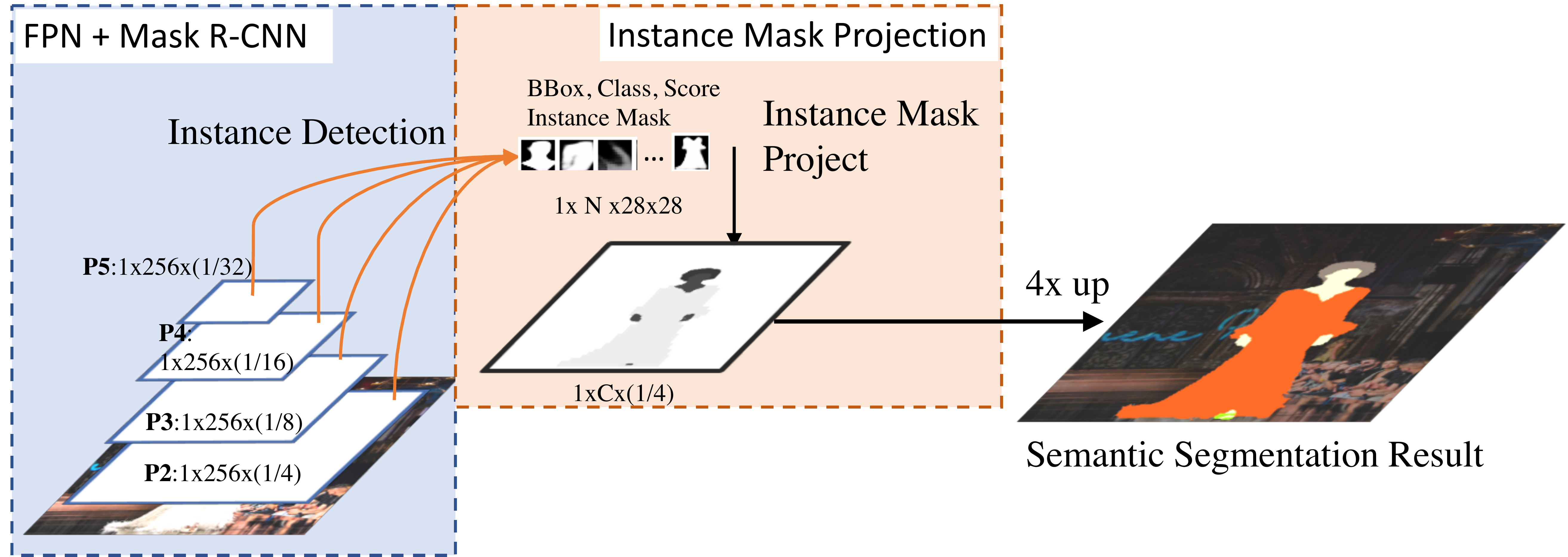}
        \caption{Mask R-CNN-IMP}
        \label{fig:arch_maskrcnn_imp}
    \end{subfigure}
    
    \begin{subfigure}{0.49\textwidth}
        \includegraphics[width=\textwidth]{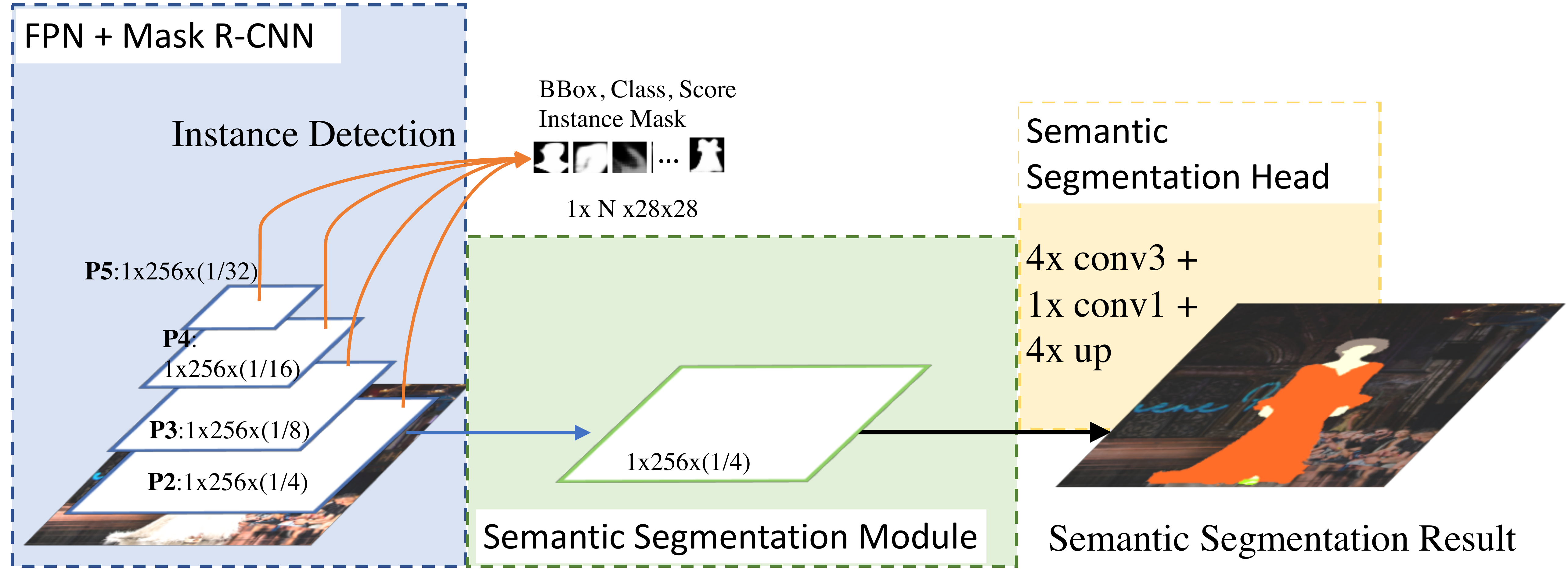}
        \caption{Panoptic-P2}
        \label{fig:arch_panoptic_p2}
    \end{subfigure}
    
    \begin{subfigure}{0.49\textwidth}
        \includegraphics[width=\textwidth]{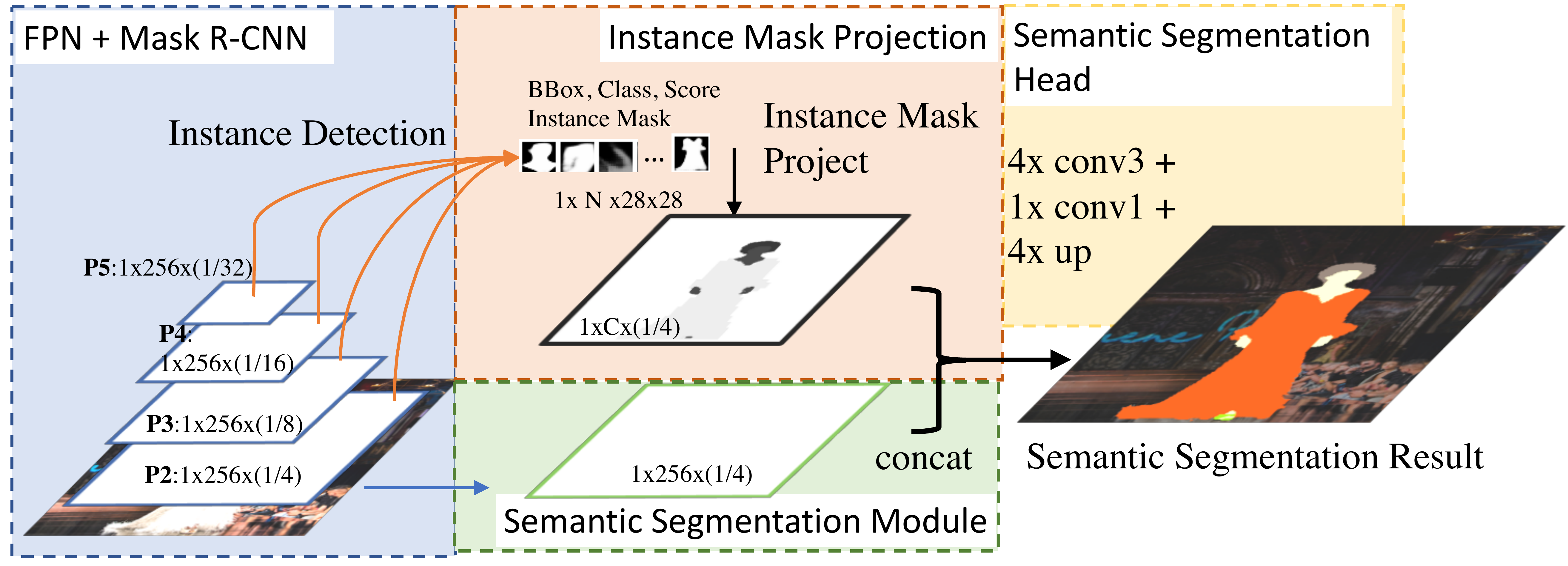}
        \caption{Panoptic-P2-IMP}
        \label{fig:arch_panoptic_p2_imp}
    \end{subfigure}
    
    \begin{subfigure}{0.49\textwidth}
        \includegraphics[height=2.5cm]{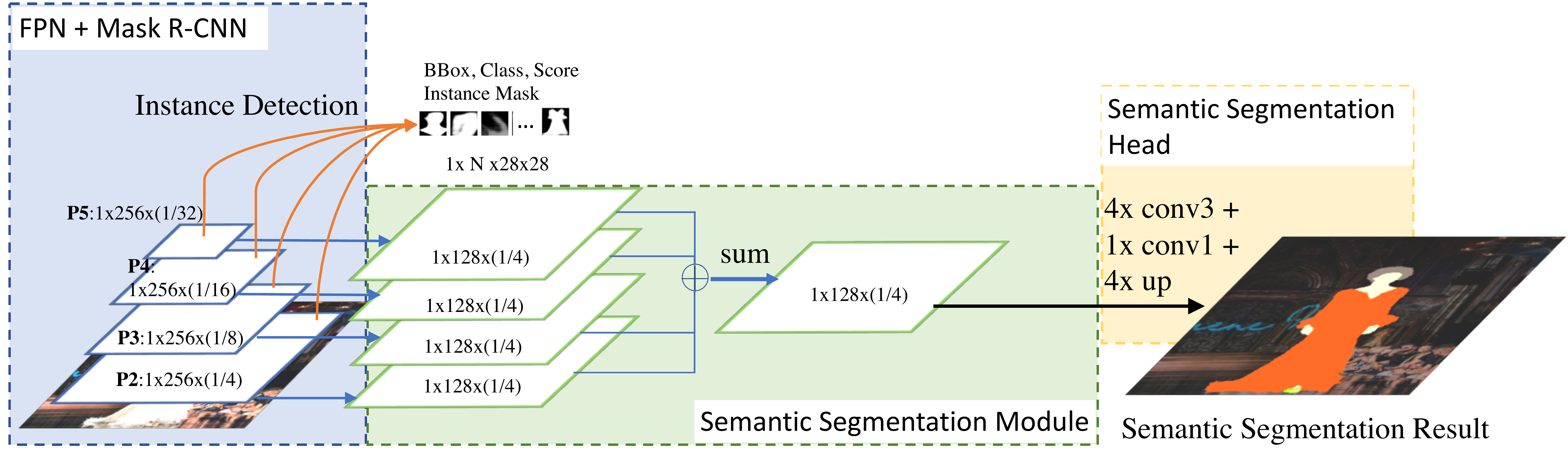}
        \caption{Panoptic-FPN}
        \label{fig:arch_panoptic_fpn}
    \end{subfigure}
    \caption{Variants of models we used in the experiments. (a) Mask R-CNN-IMP Uses the IMP to generate the semantic segmentation prediction directly without any learning parameters. (b) Panoptic-P2 uses the P2 layer in FPN to generate semantic segmentation, which is the minimal way to add semantic segmentation in FPN architecture. (c) Panoptic-P2-IMP demonstrates how to apply IMP on Panoptic-P2. (d) Panoptic-FPN combines the features layers \{P2, P3, P4, P5\} for semantic segmentation. See Figure~\ref{fig:arch_panoptic_fpn_imp} for Panoptic-FPN-IMP.}
\label{fig:arch_different}
\vspace{-3.0mm}
\end{figure}

\begin{figure*}[h]
\centering
\includegraphics[width=0.99\textwidth]{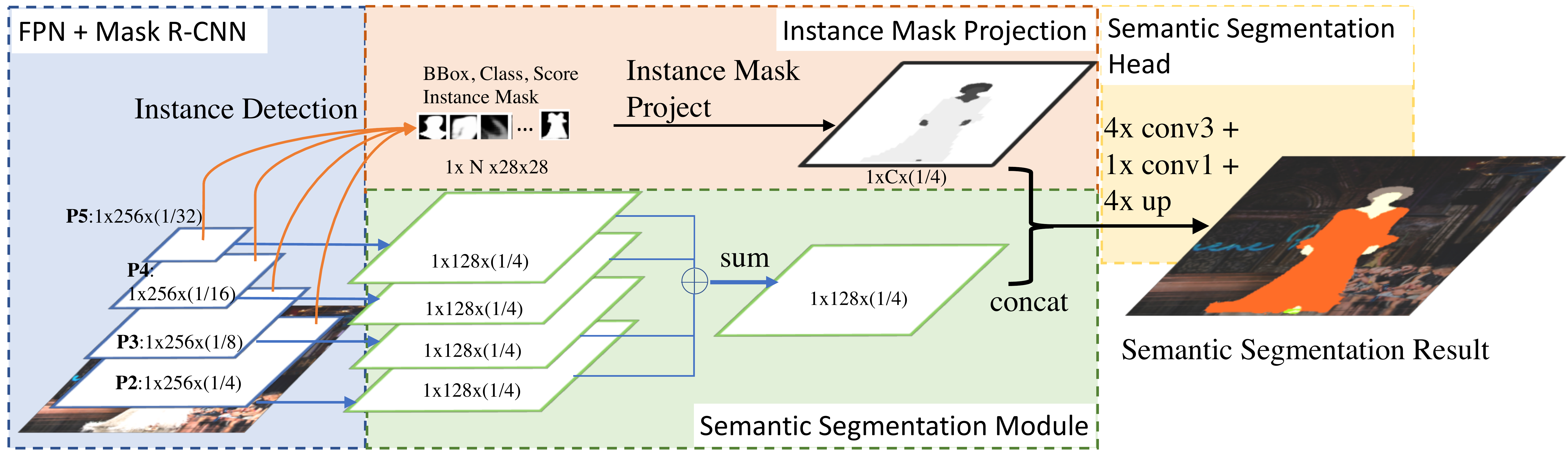}
\caption{Architecture: \textbf{Panoptic-FPN-IMP}: Our full model contains four parts. The first part is FPN + Mask R-CNN which is used for Object/Instance Detection. The Instance Mask Projection Module takes the output of instance detection to generate the feature layer(1xCx1/4). For the Semantic Segmentation Moduel, we adopts the Panoptic FPN~\cite{kirillov2019panopticfpn} which upsamples and transforms \{P2, P3, P4, P5\} to 1x128x1/4 and sums them. Then we concatenate the results of instance mask projection and semantic segmentation module and forward to the semantic segmentation prediction head. See Figure~\ref{fig:arch_different} for other models.}
\label{fig:arch_panoptic_fpn_imp}
\vspace{-3.0mm}
\end{figure*}

\vspace{-0.2cm}
\section{Model}
\vspace{-0.1cm}
Our goal is to develop a joint instance/semantic segmentation framework that can directly integrate predictions from instance segmentation to produce a more accurate semantic segmentation labeling. Our model is able to take advantage of recent advances in instance segmentation algorithms like Mask R-CNN~\cite{he2017maskrcnn} as well as advancements in semantic segmentation models~\cite{kirillov2019panopticfpn}. In this section, we first explain the proposed Instance Mask Projection (IMP) operator (Sec~\ref{sec:imp}). Next we describe how this is used to augment and improve various base models (Sec~\ref{sec:basemodels}). 


\vspace{-0.1cm}
\subsection{IMP: Instance Mask Projection}
\label{sec:imp}
\vspace{-.2cm}
The Instance Mask Projection operator projects the segmentation masks from an instance mask prediction, defined on a detection bounding box, onto a canvas defined over the whole image. This canvas is then used as an input feature layer for semantic segmentation\footnote{The resolution of the canvas can be chosen according to which feature layer is attached.}. 


 Each predicted instance mask has a Class, Score, BBox location, and $h \times w$ Mask\footnote{The resolution of Mask is 28$\times$28 in Mask R-CNN}. First the score for each pixel in the Mask is scaled by the object Score for the Class.  Then locations in the canvas layer for the Class are sampled from the scaled mask.  Note that the canvas is updated only if the scaled mask value is larger than the current canvas value. This is illustrated in Figure~\ref{fig:imp_operator} where a ``dress" is detected by Mask R-CNN and then projected onto the canvas in its detected BBox location. The projected layer shows the low resolution Instance mask which predicts outline of the dress, while the next step of semantic segmentation uses some of the FPN feature layers as well as the canvas as features and will produce a more accurate parse. 
 

This operation can be formulated as follows: 
\begin{equation*}
    \text{canvas} (c, p_{xy})  = 
    \max\left( {\text{canvas} \left(c, p_{xy}\right), S_{i}M_i\left({\text{pre}_i(p_{xy})}\right)} \right),
\end{equation*}
where there is a canvas layer for each class $c$, $p_{xy}$ is a location in the canvas, $\text{pre}_i$ maps a point in the canvas to a location in the instance mask $M_i$ for bounding box $i$, and $S_i$ is the detection score for box $i$.  Note this is only computed for $p_{xy}$ where $\text{pre}_i\left({p_{xy}}\right)$ is in the box.

This operator is applied over all detection boxes for each class independently to obtain the canvas($C\times \rfrac{H}{scale} \times \rfrac{W}{scale} $).  In the experiments the scale is 4, but this can be adjusted according to the attached feature layer.

We concatenate the IMP canvas with the feature layer(s) (either P2 or P2-5) to let the network use this as a strong prior for object location, allowing the semantic segmentation part of the model to focus on making improvements to the instance predictions during learning. 

\vspace{-0.1cm}
\subsection{Adding IMP to Base Models}
\label{sec:basemodels}
\vspace{-0.1cm}
\subsubsection*{Mask R-CNN-IMP}
\vspace{-0.2cm}
Figure~\ref{fig:arch_maskrcnn_imp} illustrates {\bf Mask R-CNN-IMP} which uses Mask R-CNN as a base model and adds IMP to project the instance masks to a canvas 
used as an approximate semantic segmentation.  This does not involve any learning or additional processing for semantic segmentation after projection and already performs well for some objects.


\vspace{-0.3cm}
\subsubsection*{Panoptic-P2, Panoptic-P2-IMP, Semantic-P2}
\vspace{-.2cm}
Next we consider lightweight versions of Panoptic FPN~\cite{kirillov2019panopticfpn} as the base model.  Panoptic FPN extends the Mask R-CNN network architecture to predict both instance segmentation and semantic segmentation.  The added semantic segmentation head takes input from multiple layers of the Feature Pyramid Network (FPN)~\cite{lin2016fpn} used in Mask R-CNN.  We perform some experiments with a lightweight version we call {\bf Panoptic-P2} that only takes features from the P2 layer of the FPN for use by the semantic prediction head (and does not use group norm) shown in Figure~\ref{fig:arch_panoptic_p2}.  When we also remove the RPN and bounding box prediction heads from {\bf Panoptic-P2}, leaving just the semantic head attached to P2 we call the network {\bf Semantic-P2}.  We experiment with adding instance mask projection to {\bf Panoptic-P2}, and call this {\bf Panoptic-P2-IMP} (illustrated in Figure~\ref{fig:arch_panoptic_p2_imp}).

\vspace{-.3cm}
\subsubsection*{Panoptic-FPN, Panoptic-FPN-IMP, Semantic-FPN}
\vspace{-.2cm}
Next, we experiment with adding IMP to the full Panoptic FPN~\cite{kirillov2019panopticfpn} calling this {\bf Panoptic-FPN-IMP}.  We also experiment with two ablated versions, {\bf Panoptic-FPN} alone (see Figure ~\ref{fig:arch_panoptic_fpn} ) and {\bf Semantic-FPN} which drops the RPN and bounding box heads from Panoptic-FPN.


Figure~\ref{fig:arch_panoptic_fpn_imp} illustrates 
Panoptic-FPN-IMP which uses the conv3x3(128) + GroupNorm~\cite{wu2018gn} + ReLU + Bilinear upsampling(2x) as the upsampling stage. 
For P3(scale/8), P4(scale/16), P5(scale/32) layers, we first upsample each to (1/4) scale. For the P2 layer, we apply conv3x3 to reduce the dimension from 256 to 128. Then, we sum these 4 layers to ($128 \times \rfrac{H}{4} \times \rfrac{W}{4}$) and concatenate with the Instance Mask projected layer to form the feature layer((128 + $C) \times \rfrac{H}{4} \times \rfrac{W}{4}$). Finally, we apply 4 conv3x3 and 1 conv1x1 layers to generate semantic segmentation predictions. In contrast to FPN-P2, all conv3x3 use Group Norm.  




\vspace{-0.2cm}
\subsection{Training}
\label{sec:training}
\vspace{-0.1cm}
We adopt a two stage training solution, first training a Mask R-CNN detection/instance segmentation model then using this as an initial prediction for training our full model. Pre-training is incorporated for practical reasons to reduce training time (without pre-training the IMP will vary significantly over training iterations, making convergence slow).
In the first stage, we follow the Mask R-CNN training settings but adjust the parameters for 4 GPU machines (Nvidia 1080~Ti) by following the Linear Scaling Rule~\cite{goyal2017imagenet1hr}. For implementation we use PyTorch v1.0.0~\cite{paszke2017pytorch} and base our code on the maskrcnn-benchmark repository~\cite{massa2018mrcnn}.

\vspace{-0.2cm}
\section{Experiments}
\label{experiments}
\vspace{-0.1cm}
We evaluate our proposed model on two different tasks: clothing parsing and street scene segmentation.

\label{experiments}

\vspace{-0.1cm}
\subsection{\Shopagon}
\label{sec:experimentsVCD}
\vspace{-0.1cm}
The \Shopagon~is for clothing parsing -- where the goal is to assign an apparel category label (e.g. shirt, skirt, sweater, coat, etc) to each pixel in a picture containing clothing. This is an extremely challenging segmentation problem due to clothing deformations and occlusions due to layering.
The dataset depicts 25 clothing categories, plus skin, hair, and background labels, with pixel-accurate polygon segmentations, hand labeled on 6k images. 
The dataset covers a wide range of depictions, including: real-world pictures of people, layflat images (clothing items arranged on a flat surface), fashion-runway photos, and movie stills. Special care is taken to sample clothing photos from around the world, across varied body shapes, in widely varied poses, and with full or partial-bodies visible.

Since this dataset was initially collected for clothing parsing, a single garment may be split into multiple segments (e.g. a shirt worn under a buttoned blazer may appear as a segment at the neck, plus 2 shirt cuff segments at each wrist). To convert the semantic segmentations into instance annotations, each segment (connected component) is treated as an instance with corresponding bounding box. This definition is slightly different than COCO~\cite{lin2014coco} or Cityscapes~\cite{Cordts2016Cityscapes} and produces more small instances. However, we experimentally observe benefits to this approach over combining all segments from a garment into a single instance/BBox because it doesn't require the model to make long range predictions across large occlusions.   

In our experiments, the train and validation sets contain 5493 and 500 images respectively and all images are ~1280$\times$720 pixels or higher. 
For training the first stage, we use an ImageNet Classification pre-trained model, with prediction layer weights initialized according to a normal distribution(mean=0, standard derivation=0.01). We set batch size to 8, learning rate to 0.01, and train for 70,000 iterations, dropping the learning rate by 0.1 at 40,000 and 60,000 iterations. We also use this setting for training the second stage (including the semantic segmentation branch). For the input image, we resize the short side to 800 pixels and limit the long side to 1333.

\begin{table}[t]
\small
    \centering
    \begin{tabular}{c|l|c|c|c| c}
         & \multirow{2}{*}{Model} & \multirow{2}{*}{BBox} & \multirow{2}{*}{Mask} &  \multicolumn{2}{c}{Semantic} \\
          &     &    &   & mIOU & mAcc \\     
         \hline
         \hline
         1 & Mask R-CNN& 29.9 & 26.7 & NA & NA\\
         2 & Mask R-CNN-IMP & & & 43.91 & 56.93 \\
        \hline 
        3 & Semantic-P2 & NA & NA & 37.00 & 48.57 \\
        4 & Semantic-FPN & NA & NA &  42.66 & 55.19\\
        
        \hline 
        5 & Panoptic-P2 & 29.8 &  26.4 &  37.14 & 48.82 \\
        6 & Panoptic-P2-IMP & \textbf{30.6} & \textbf{26.8} & 46.59 & 59.24 \\
        7 & Panoptic-FPN & 29.6 & 26.7 & 45.01 & 57.08 \\
        8 & Panoptic-FPN-IMP & 30.4 & \textbf{26.8} & \textbf{47.03} & \textbf{61.52} \\
        
    \end{tabular}
    \caption{Ablation Study on~\Shopagon. The backbone network is ResNet-50. We train the model with different settings, Panoptic-P2 v.s. Panoptic-FPN, w/wo Instance Mask Projection(IMP), w/wo BBox/Mask prediction head. For the BBox, and Mask, we use the COCO evaluation metric. For the semantic segmentation metric, we use meanIOU and mean Accuracy.}
    \label{tab:shoapgon_ablation_study}
     \vspace{-2.0mm}
\end{table}

Table~\ref{tab:shoapgon_ablation_study} shows the performance of our models under different settings with ResNet-50 as the backbone network. 
First, we report the performance of baseline instance (row 1) and semantic segmentation models (rows 3-4). Next, we show results on Panoptic models that integrate instance and semantic segmentation (Panoptic-P2 and Panoptic-FPN, rows 5 and 7). Adding our proposed IMP operator significantly increases semantic segmentation performance when incorporated into each of these base models (rows 6 and 8), improving absolute performance of Panoptic-P2 by 9.45 mIOU and 1.42 in mAcc, and improving Panoptic-FPN by 2.02 mIOU and 4.44 in mAcc. For reference, we also experiment with adding IMP to the base Mask R-CNN model (row 2), and achieve semantic segmentation performance better than Semantic-FPN and Panoptic-P2, and comparable to Panoptic-FPN without requiring any dedicated semantic segmentation branch.



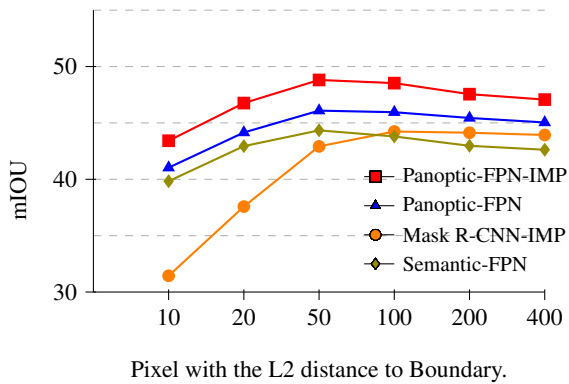
\begin{figure}[th]
\begin{tikzpicture}[y=.15cm, x=1cm]
    \small
        \draw[black!30, dashed] (0, 5) -- (6, 5);
    \draw[black!30, dashed] (0, 10) -- (6, 10);
    \draw[black!30, dashed] (0, 15) -- (6, 15);
    \draw[black!30, dashed] (0, 20) -- (6, 20);
    \draw[black!30, dashed] (0, 25) -- (6, 25);

	\draw (0,0) -- coordinate (x axis mid) (6, 0);
    \draw (0,0) -- coordinate (y axis mid) (0,25);
    \foreach \x\xtext in {1/{10}, 2/{20}, 3/{50}, 4/{100}, 5/{200}, 6/{400}}
     	\draw (\x,1pt) -- (\x,-3pt)
			node[anchor=north] {\xtext};
    \foreach \y\ytext in {0/30, 10/40,  20/50}
     	\draw (1pt,\y) -- (-3pt,\y) 
     		node[anchor=east] {\ytext}; 
	\node[below=0.8cm] at (x axis mid) {Pixel with the L2 distance to Boundary.};
	\node[rotate=90, above=0.8cm, left=1cm] at (y axis mid) {mIOU};

	\draw[red, thick] plot[mark=square*, mark options={fill=red}] 
		file{fpn-panoptic-imp};
	\draw[orange, thick] plot[mark=*, mark options={fill=orange} ] 
		file{maskr-cnn-imp};
	\draw[blue, thick] plot[mark=triangle*, mark options={fill=blue}] 
		file{maskr-cnn-semantic};
    \draw[olive, thick] plot[mark=diamond*, mark options={fill=olive}]
        file{maskr-cnn-semantic-only};
    
	legend
	\begin{scope}[shift={(3.6,2.5)}] 
	
	\draw (0,0) -- 
		plot[mark=diamond*, mark options={fill=olive}] (0.15,0) -- (0.4,0)
		node[right]{\footnotesize{Semantic-FPN}};
	\draw [yshift=\baselineskip] (0, 0) -- 
		plot[mark=*, mark options={fill=orange}] (0.15,0) -- (0.4,0)
		node[right]{\footnotesize{Mask R-CNN-IMP}};
		
	\draw [yshift=2\baselineskip](0,0) -- 
		plot[mark=triangle*, mark options={fill=blue}] (0.15,0) -- (0.4,0)
		node[right]{\footnotesize{Panoptic-FPN}};
	\draw[yshift=3\baselineskip]  (0,0) -- 
    		plot[mark=square*, mark options={fill=red}] (0.15,0) -- (0.4,0) 
		node[right]{\footnotesize{Panoptic-FPN-IMP}};

	\end{scope}
\end{tikzpicture}
\caption{Analysis of performance of pixel within the distance to the boundary. In this Figure, we adopt the Panoptic FPN as the backbone network and shows 4 models, Semantic-FPN, Mask R-CNN-IMP, Panoptic-FPN, and Panoptic-FPN-IMP.}
\label{fig:shopagon_distance_analysis}
\vspace{-2.0mm}
\end{figure}

\begin{table*}[bt]
\centering
\footnotesize
 \resizebox{\linewidth}{!}{
    \begin{tabular}{l|c|c|c|c|c|c|c|c|c|c|c|c|c|c}
        \multirow{2}{*}{Model} & \multirow{2}{*}{mean} &  \multirow{2}{*}{bag} & \multirow{2}{*}{belt} & \multirow{2}{*}{boots} & \multirow{2}{*}{foot-} & \multirow{2}{*}{outer} & \multirow{2}{*}{dress} & \multirow{2}{*}{sun-} & \multirow{2}{*}{pants} & \multirow{2}{*}{top} & \multirow{2}{*}{shorts} & \multirow{2}{*}{skirts} & \multirow{2}{*}{head} & \multirow{2}{*}{scarf\&} \\
        & & & & & wear & & & glasses& & & & & wear & tie \\
        \hline 
        
        FCN-32~\cite{Shelhamer2016FCN}  &35 &27&12&32&33&36&28&25&51&38&40&28&33&17\\

        FCN-16~\cite{Shelhamer2016FCN} &
        37 & 26& 19& 32& 38 & 35& 25& 37& 51& 38& 40& 23& 41& 16 \\
        
        FCN-8~\cite{Shelhamer2016FCN} & 38  & 24 & 21 & 32 & 40 & 35 & 28 & 41 & 51 & 38 & 40 & 24 & 44 & 18 \\
        
        FCN-8satonce~\cite{Shelhamer2016FCN} & 38 & 26 & 20 & 31 & 40 & 35 & 29 & 36 & 50 & 39 & 38 & 26 & 44 & 16 \\

        CRFasRNN~\cite{zheng2015crfasrnn} & 41 & 30 & 18 & 41 & 39 & 43 & 32 & 36 & 56 & 40 & 44 & 26 & 45 & 22 \\
        DeepLabV3+~\cite{chen2018depthwise_atrous_conv} & 51 & 42 & 28 & 40 & 51 & 56 & 52 & 46 & 68 & 55 & 53 & 41 & 55 & 31 \\
        \hline
        Ours: & & & & & & & & &  & & & & \\
        \textbf{R50 Panoptic-P2-IMP} & 69.7 &  74.8  & 57.4 & \textbf{59.7} & \textbf{59.4} & 69.2 & 64.2 & 68.5 & 77.2 & 67.7 & 71.9 & 62.7 & 75.3 & 97.5 \\

        \textbf{\footnotesize{R50 Panoptic-FPN-IMP}} & 71.1 & 77.1 & 58.1 &  57.9 & 59.1 & \textbf{72.2} & 68.2 & 68.4 & \textbf{80.4} & 68.7 & 72.5 & \textbf{67.9} & 76.2 & \textbf{97.9}  \\
         
        \textbf{\footnotesize{R101 Panoptic-FPN-IMP}} & 
        \textbf{71.4} & 
        \textbf{77.9} &  \textbf{59.0}  & 58.8 & \textbf{59.4} & 72.0 & \textbf{68.3} & \textbf{68.6} & 79.3 & \textbf{69.1} & \textbf{74.1} & 67.8 & \textbf{76.4} & \textbf{97.9}\\

    \end{tabular}
    }
    \caption{Comparison to the baseline models provided by ModaNet on IOU metric. Our model shows 20.4\% absolutely improvement for mean IOU. For certain categories, especially those whose size is quite small such as belt, sunglasses, headwear and scarf \& tie, our models show dramatic improvement. For simplicity, we use R50 and R101 to represent ResNet0-50 and ResNet-101.}  
    \label{tab:modanet_sota}
     \vspace{-2.0mm}
\end{table*}

Another question we consider is how much this method helps refining object boundaries, since producing an accurate object contour may be necessary for applications like visual search or virtual clothing try-on. In Figure~\ref{fig:shopagon_distance_analysis}, we analyze the mIOU of pixels within 10-400 L2 distance from the boundary. Generally, we observe that for pixels close to the boundary, semantic and instance/semantic methods all perform much better than Mask R-CNN-IMP and this gap decreases for larger distances. This is because Mask R-CNN generates 28$\times$28 instance masks. Therefore, once we project the instance segmentation results on the canvas, the boundary will not be sharp, but pixels near the center of the object will be labeled correctly. We also generally observe larger improvements of the IMP operator on pixels near the boundary, with benefits dropping off for central pixels.


In Figure~\ref{fig:vis_cases}, we show some qualitative examples. In some cases,~\ref{fig:vis_woman_in_dress},~\ref{fig:vis_man_wears_tie}, Mask R-CNN-IMP already produces a better semantic segmentation than the Panoptic-FPN architecture. We also observe that often, when an object is small (tie, watch), or plain and covering a large area, IMP enhanced methods generally perform better. In Figure~\ref{fig:vis_girl_in_jacket}, by combining the semantic segmentation features and IMP, our model fixes category confusions occuring on different regions of an object. Although most training images in the \Shopagon  only contain one person per image, we see that our model generalizes well to complicated examples containing multiple people (Figure~\ref{fig:vis_two}).

\vspace{-0.1cm}
\subsection{ModaNet}
\label{sec:experimentsModaNet}
\vspace{-0.1cm}

 ModaNet~\cite{zheng2018modanet} is a large clothing parsing dataset, containing annotations for BBox, instance-level masks, and semantic segmentations. It contains ~55k images (52,377 images in training and 2,799 images in validation), sampled on an existing fashion focused dataset of images from the Chictopia website. The ModaNet data is relatively low resolution (640x480 or smaller) compared to the \Shopagon~ data, sampled to generally contain a single full-body depiction of a standing person, centrally located in the image. 13 clothing categories are labeled (without skin, hair, or background) at relatively high fidelity (but less pixel-accuracy than the \Shopagon). 


We use a similar two-stage ImageNet classification pre-training method as for the \Shopagon, training for 90k iterations, dropping the learning rate at 60k and 80k iterations. Here, we
resize the input image to limit its short side to 600 and long side to 1000. During training, we use multi-scale training by randomly changing the short side to \{400, 500, 600, 700, 800\}.

\begin{table}[h]
    \centering
    \begin{tabular}{l|c|c|c}
         Model & BBox & Mask & Semantic\\
               &      &      &  (mIOU) \\
         \hline
         Semantic-P2 & NA & NA & 64.60 \\
         Panoptic-P2 & 57.2 & 55.5 & 65.93 \\
         Mask R-CNN-IMP & 57.2 & 55.5 & 66.23 \\
         Panoptic-P2-IMP & \textbf{58.0} & \textbf{55.9} & 69.65 \\
         \hline 
         Panoptic-FPN-IMP & 57.8 &  55.6  & \textbf{71.41}\\
        
    \end{tabular}
    \caption{Results on ModaNet with ResNet-50 as the backbone model. Panoptic-P2-IMP and Mask R-CNN-IMP both provide improvements on semantic segmentation compared to Semantic-P2 and Panoptic-P2. 
    }
    \label{tab:modanet_ablation}
    \vspace{-5.0mm}
\end{table}

Table~\ref{tab:modanet_ablation} shows experimental results demonstrating the addition of the IMP operator. We evaluate baseline models, Semantic-P2 and  Panoptic-P2, 64.60\% and 65.93\% mIOU, respectively. Compared to these models, we see that Mask R-CNN-IMP can generate better results on semantic segmentation without a dedicated semantic segmentation head. This also matches our previous experiments on the \Shopagon. Adding IMP to Panoptic-P2, Panoptic-P2-IMP achieves a semantic performance of 69.65\%, outperforming Panoptic-P2 by 3.72\% mIOU and Panoptic-FPN-IMP even further improves mIOU to 71.41\%.


In Table~\ref{tab:modanet_sota}, we also train our final model, Panoptic-FPN-IMP with ResNet-101 and compare to the baseline results provided by ModaNet~\cite{zheng2018modanet}. First, our model achieves 20.4\% absolutely mIOU improvement compared to the best performing semantic segmentation algorithm, DeepLabV3+, provided by ModaNet. Plus, we achieve more consistent results, scoring over 50\% IOU for each class.
Compared to the baseline results, our model does extremely well on small objects, e.g. belt, sunglasses, headwear, scarf\&tie (on scarf\&tie we achieve 97.9\% mIOU). 
We have some speculations about these improvements. Compared to semantic segmentation methods which tend to base their predictions on fixed scale local regions, object detection takes context from the dynamically chosen region around the object, providing an advantage for segmentation. We also observe improvements on confusing classes, e.g. the bottom part of a dress is visually similar to a skirt. Purely semantic segmentation methods may not be able to differentiate ambiguous cases as well as methods that exploit context determined by object detection.


\begin{table*}[th]
\small
\centering

    \begin{tabularx}{1.0\textwidth}{l|X|X|X|X|X|X|X|X|X|X|X||X|X|X|X|X|X|X|X }
    Type & \multicolumn{10}{c}{Stuff class} & &\multicolumn{7}{c}{Things class}  \\
    \hline
    {\footnotesize Model} & {\scriptsize road} & {\scriptsize side-walk} & {\scriptsize build-ing} & {\scriptsize wall} & {\scriptsize fence} & {\scriptsize  pole}  & {\scriptsize traffic light} & {\scriptsize traffic sign} & {\scriptsize vegeta-tion} & {\scriptsize terrain} & {\scriptsize sky} & {\scriptsize person} & {\scriptsize rider} & {\scriptsize car} & {\scriptsize truck} & {\scriptsize bus} & {\scriptsize train} & {\scriptsize motor-cycle} & {\scriptsize bicycle}  \\
    \hline

    \multicolumn{6}{l|}{\textit{Without all the Data Augmentation}} & & & &  & & &&&& & & & \\
     & 97.7 & 81.7 & 91.2 & 41.2 & 51.7 & 58.8 & 67.3 & 74.6 & 91.6 & 59.3 & 93.8 & 81.2 & 60.3 & 93.6 & 61.4 & 80.4 & 63.2 & 57.0 & 76.1 \\
    IMP & 97.6 & 81.5 & 91.2 & 39.6 & 52.0 & 59.2 & 66.6 & 74.9 & 91.5 & 59.7 & 93.8 & 81.9 & 64.7 & 93.8 & 63.9 & 81.6 & 74.0 & 63.5 & 76.7 \\
    
    \hline 
    \multicolumn{6}{l|}{\textit{With all the Data Augmentation}} & & & &  & & &&&& & & & \\
     & 97.7 & 82.5 & 91.7 & 45.0 & 56.4 & 61.4 & 69.6 & 77.1 & 91.7 & 60.1 & 94.3 & 82.4 & 64.0 & 94.7 & 74.5 & 84.5 & 77.6 & 62.9 & 77.9 \\
    
    IMP & 97.9 & 83.6 & 91.4 & 38.3 &  55.9  & 62.0 & 69.9 & 77.5 & 91.9 & 59.8 & 94.5 & 83.5 & 69.1 & 95.1 & 83.9 & 91.4 & 83.1 & 67.2 & 78.7 \\
    
    \end{tabularx}
    
    \caption{Comparisons of per Class IOU with and without IMP on Cityscapes. We show two scenarios without (top) and with (bottom) data augmentation. We see Instance Mask Projection(IMP) improves both scenarios. For Thing classes, we see 4.2/3.2 mIOU improvement with/without all data augmentation.}
    \label{tab:cityscape_perclass_iou}
    \vspace{-2.0mm}

\end{table*}

\vspace{-0.1cm}
\subsection{Cityscapes}
\label{sec:experimentsCityscapes}
\vspace{-0.1cm}

We also experiment on Cityscapes~\cite{Cordts2016Cityscapes}, an ego-centric self-driving car dataset. All images are high-resolution (1024$\times$2048) with 19 semantic segmentation classes, and instance-level masks for 8 thing-type categories. 
The collection contains two sets, fine-annotation and coarse-annotation sets. We focus our experiments on fine-annotation, containing 2975/500/1525 train/val/test images. 

For Cityscapes, we use the COCO model as the pre-trained model, reusing the weights in the prediction layer for all classes except ``Rider" which does not exist in COCO (weights are randomly initialized). Then, the input is resized to  1024$\times$2048 , or 800$\times$1600 randomly. We follow Panoptic FPN~\cite{kirillov2019panopticfpn} to add three data augmentations: {\em multi-scaling}, {\em color distortion}, and {\em hard boostrapping}. For multi-scaling, the short side of the input image is resized to \{512, 724, 1024, 1448, 2048\} randomly and cropped to 512$\times$1024. The color distortion randomly increases/decreases brightness, contrast, and saturation 40\%, and shifts the Hue \{-0.4, 0.4\}. Hard boostrapping selects the top 10, 25, 50 percent of pixels for the loss function. 
In contrast to ~\Shopagon~ and ModaNet, we skip the first-stage training, since the pretrained model from COCO already provides strong enough performance. We set batch size to 16, learning rate to 0.005, and train for 130,000 iterations, dropping the learning rate by 0.1 at 80,000 and 110,000 iterations.


For Cityscapes, we focus evaluations on the FPN-Panoptic network (ablation study in Table~\ref{tab:cityscape_ablation}). Model(a) is the Mask R-CNN model. Model(b) is the Panoptic-FPN model without data augmentation. For ColorJitter, model(b) and (d) are the comparison set (improvement from ColorJitter is not clear). In model(d) to model(h), Multi-scale training definitely helps a lot and  also reduces overfitting on BBox/Mask prediction. For hard bootstrapping, we see consistent improvements when setting the lower ratio from Model(e), Model(f), to Model(g). Instance Mask Projection provides around 1.35/1.5 improvement in Model(b) to Model(c) without any data augmentation and Model(i) to Model(j) with all data augmentations. 

Compared to the \Shopagon~and ModaNet, we observe less dramatic overall improvement from IMP. However, one reason is that only 8 of 19 classes are "thing" like categories where we expect our method to be most helpful. In Table~\ref{tab:cityscape_perclass_iou}, we show two comparison sets (with and without data augmentation) for each Cityscapes class.
For the Stuff classes, the difference are minor, except `Wall` (-1.6/-6.7). For the Thing classes, certain classes are improved dramatically, especially those that have fewer training instances or that are smaller, i.e. Rider, Truck, Bus, Train, Motorcycle. In fact, over all Thing classes we observe a mIOU increase of 4.2/3.2, with and without data augmentation respectively.  


Besides ResNet-50, we also train our final model, Panoptic-FPN-IMP with ResNet-101 and ResNeXt-101-FPN to compare with state-of-the-art methods on Cityscapes \texttt{val} set (Table~\ref{tab:cityscape_state-of-the-art}). Our method is still better than Panoptic FPN~\cite{kirillov2019panopticfpn}, though the improvements are reduced when using more complex models. 
We still see our simple model can achieve similar performance to those models using heavily engineering methods.

\begin{table}[ht]
\small
    \centering
    \resizebox{0.95\columnwidth}{!}{%
    \begin{tabular}{l|l|c|c|c|c|c|c|c}
         & Model & \footnotesize{Color} & \footnotesize{MS} & BS & IMP & Box & Mask & mIOU\\
         \hline
         a & \multicolumn{2}{l|}{\footnotesize{Mask R-CNN}} & & & & 40.9 & 35.5 & NA\\ 
         \hline 
         b & \multicolumn{2}{l|}{\footnotesize{Panoptic-FPN}}  & & & &  36.9 & 32.7 & 72.74 \\
         c&  &  & & & \ding{52} & 36.9 & 32.5 & 74.09  \\
         d& & \ding{52} &  & & & 36.8 & 32.8 & 73.12  \\
         e& & \ding{52} &  & 0.50 & & 37.8 & 34.0 & 73.81 \\
         f& & \ding{52} &  & 0.25 & & 38.4 & 34.1 & 73.93 \\ 
         g& & \ding{52} &  & 0.10 & & 38.7 & 34.7 & 74.94 \\ 
         h& & \ding{52} & \ding{52} & & & 39.9 & 35.9 &  75.99 \\
         i& & \ding{52} & \ding{52} & 0.10 & &  40.7 & 36.5 & 76.11 \\
         j& & \ding{52} & \ding{52} & 0.10 & \ding{52} & 39.8 & 35.8  & 77.49 \\
          
         & 
    \end{tabular}
    }
    \caption{Performance Analysis of each module used on Cityscapes \texttt{val} set. For simplicity, we use the following abbreviation: \textbf{MS}:\textit{multi-scale training}, \textbf{Color}:\textit{Color Jitter},
    \textbf{BS}:\textit{Hard Boostraping},
    \textbf{IMP}:\textit{Instance Mask Projection},
    }
    \label{tab:cityscape_ablation}
    \vspace{-2.0mm}
\end{table}

\begin{table}[h]
    \centering
    \begin{tabular}{l|l|c}
        Method & Backbone & mIOU \\
        \hline 
        
        PSANet101 ~\cite{zhao2018psanet} & ResNet-101-D8 & 77.9 \\
        Mapillary~\cite{rotabulo2018place} &  WideResNet-38-D8 & 79.4 \\
        DeeplabV3+ ~\cite{chen2018depthwise_atrous_conv} & X-71-D16 & 79.6 \\
        \hline \hline
        Panoptic FPN~\cite{kirillov2019panopticfpn} & ResNet-101-FPN  & 77.7 \\
         & ResNeXt-101-FPN & 79.1 \\
        \hline
        Ours:Panoptic-FPN-IMP & ResNet-50-FPN & 77.5\\
         & ResNet-101-FPN & 78.3 \\
         & ResNeXt-101-FPN & 79.4\\
    \end{tabular}
    \caption{Comparisons on Cityscapes \texttt{val} set. Our models obtain 0.6 and 0.3 mIOU improvement over Panoptic-FPN~\cite{kirillov2019panopticfpn} on the same backbone architectures.}
    \label{tab:cityscape_state-of-the-art}
     \vspace{-2.0mm}
\end{table}

\begin{table}[h]
    \centering
    \footnotesize
    \begin{tabular}{c|c|l|r}
         Resolution & Backbone  & Model  &  Speed(ms) \\
         \hline \hline
         \multicolumn{2}{c|}{\Shopagon}  \\
         \hline 
          \multirow{5}{*}{800$\times$ 1333} & \multirow{5}{*}{R50} &   Mask R-CNN & 92\\
         & & Mask R-CNN-IMP & 94 \\
         & & Semantic-FPN & 103\\
         &  &Panoptic-FPN & 110  \\
         &  &Panoptic-FPN-IMP & 111  \\
        \hline\hline
        \multicolumn{2}{c|}{ModaNet}  \\
        \hline
        \multirow{2}{*}{600$\times$1000} & R50 &  Panoptic-FPN-IMP & 72 \\
        \cline{2-4}
        & R101 &  Panoptic-FPN-IMP & 87 \\
         
        \hline\hline         
        \multicolumn{2}{c|}{Cityscapes}  \\
        
        \hline
         \multirow{5}{*}{1024$\times$2048}  & R50   & Mask R-CNN & 151\\
         & & Panoptic-FPN & 194 \\  
         & & Panoptic-FPN-IMP & 195\\       
          \cline{2-4}
         & R101   &Panoptic-FPN-IMP & 243 \\
          \cline{2-4}
         & X101  &Panoptic-FPN-IMP &
         401 \\
    \end{tabular}
    \caption{Speed performance analysis. In this table, we show the speed performance for each model. For simplicity, we use the following abbreviations:\textbf{R50}:ResNet-50. \textbf{R101}:ResNet-101. \textbf{X101}:ResNeXt-101}
    \label{tab:time_analysis}
    \vspace{-0.2cm}
\end{table}

\vspace{-0.2cm}
\subsection{Inference Speed Analysis}
\vspace{-0.1cm}

Table~\ref{tab:time_analysis}, shows some speed performance analysis for each dataset.
Due to the different number of instance classes and input resolutions, the speed performance of models can vary. In experiments, we find the results are quite consistent and very efficient, adding IMP only costs 1$\sim$2 ms in inference on top of each baseline model. In all experiments, the result is from a single output without any bells and whistle.

\vspace{-0.2cm}
\section{Conclusion}
\vspace{-0.2cm}
In this work, we propose a new operator, Instance Mask Projection, which projects the results of instance segmentation as a feature representation for semantic segmentation. This operator is simple but powerful. Experiments adding IMP to Panoptic-P2/Panotpic-FPN show consistent improvements, with negligible increases in inference time. Although we only apply it on the Panoptic-P2/Panoptic-FPN, this operator can generally be applied to other architectures as well. 


\section*{Appendix}

\subsection*{~\Shopagon~Classes}

\begin{table}[h]
    \centering
    \resizebox{\linewidth}{!}{
    \begin{tabular}{l|l|r|r|r}
    Class & Super Class  & $\#$ Train & $\#$ Val & Area($x^2$)\\
    \hline
    Hair & Body  & 7,260 & 635 & 192 \\ 
    Skin & Body  & 34,795 & 3,074 & 119\\ 

    Top/T-shirt & G-Top  & 4,364 & 424 & 221\\
    Sweater/Cardigan & G-Top  & 1,906 & 148 & 266 \\
    Jacket/Blazer & G-Top  & 2,360 & 183 & 261\\
    Coat & G-Top & 1,597 & 161 & 279\\
    Shirt/Blouse & G-Top & 2,650 & 244 & 229\\
    Vest & G-Top & 266 & 20 & 220\\

    Pants/Jeans & G-Bottom  & 2,763 & 217 & 261\\
    Tights/Leggings & G-Bottom & 930 & 116 & 214\\
    Shorts & G-Bottom  & 532 & 60 & 203\\
    Socks & G-Bottom & 803 & 80 & 174 \\
    Skirt & G-Bottom & 1,281 & 114 & 262 \\ 
    
    Dress & G-Whole & 2,728& 241 & 340\\
    Jumpsuit & G-Whole & 273 & 31 & 370\\ 
    
    Shoes & Footwear & 6,619 & 591 & 118 \\
    Boots & Footwear & 1,801 & 109 & 142\\
    
    Hat/Headband & Accessories & 983 & 111 & 192\\
    Scarf/Tie & Accessories & 909 & 88 & 274\\ 
    Watch/Bracelet & Accessories & 2,627 & 206 & 86\\
    Bag & Accessories & 3,284 & 263 & 186 \\ 
    Gloves & Accessories & 431 & 41 & 210\\ 
    Necklace & Accessories & 1,711 & 134 & 131\\
    Glasses & Accessories & 1,329 & 129 & 89\\ 
    Belt & Accessories & 1,035 & 95 & 110 \\ 
    \end{tabular}
    }
    \caption{\Shopagon~Class Definition and statistics.}
    \label{fig:shopagon_statistics}
\end{table}

Table~\ref{fig:shopagon_statistics} shows the class definition and statistics of the ~\Shopagon. Because we convert each segment(connected component) of semantic segmentation into an instance annotation, the number of training instance is much more than usual. The details can be found in Sec.~\ref{sec:experimentsVCD} in the main submission. Another is the diverse classes. In contrast to ModaNet~\cite{zheng2018modanet}, in ~\Shopagon, the confusing classes are not grouped. For example, Jacket/Blazer to Coat. This makes it more challenging for semantic segmentation approaches to generate clean results.

\begin{figure*}[th!]

    \begin{subfigure}{\textwidth}
        \centering
        \begin{subfigure}[b]{0.241\textwidth}
            \caption*{Image}
            \includegraphics[width=\textwidth]{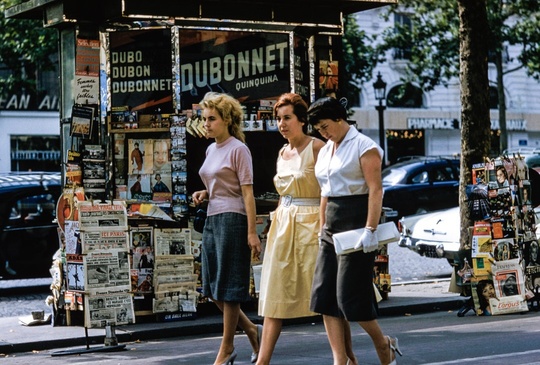} 
        \end{subfigure}
        \hfill
        \begin{subfigure}[b]{0.241\textwidth}
            \caption*{Panoptic-FPN}
            \includegraphics[width=\textwidth]{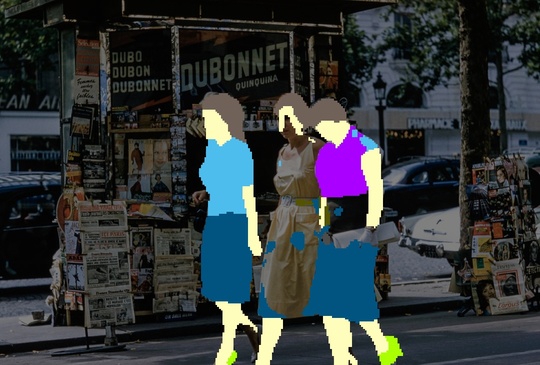}
        \end{subfigure}
        \hfill
        \begin{subfigure}[b]{0.241\textwidth}
            \caption*{ Mask R-CNN-IMP}
            \includegraphics[width=\textwidth]{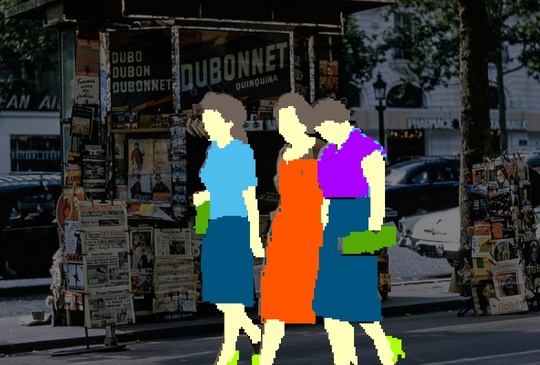}
        \end{subfigure}
        \hfill
        \begin{subfigure}[b]{0.241\textwidth}
            \caption*{Panoptic-FPN-IMP}
            \includegraphics[width=\textwidth]{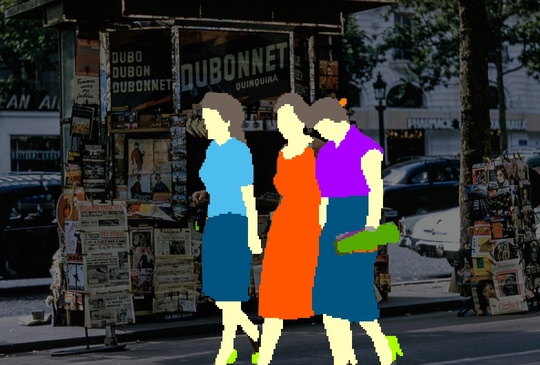}
        \end{subfigure}
    \end{subfigure}
    
    \begin{subfigure}{\textwidth}
        \includegraphics[width=0.241\textwidth]{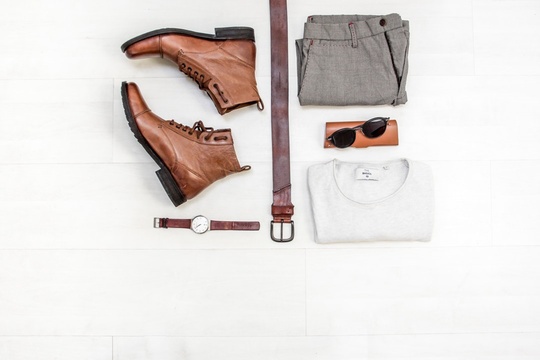}
        \hfill
        \includegraphics[width=0.241\textwidth]{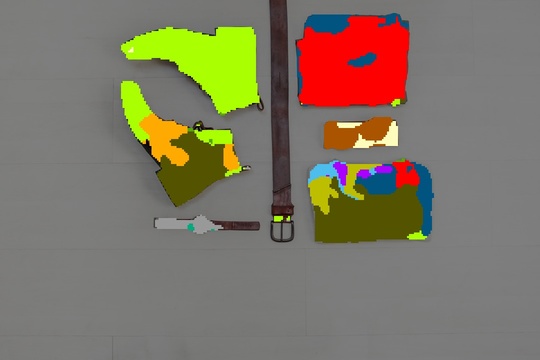}
        \hfill
        \includegraphics[width=0.241\textwidth]{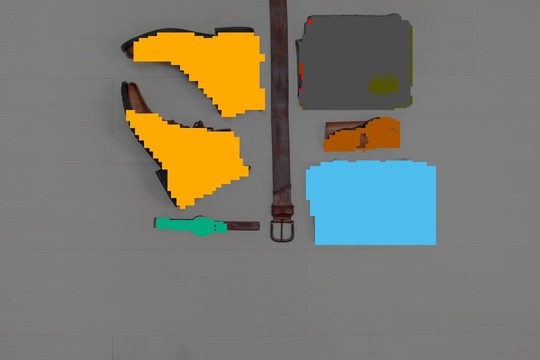}
        \hfill
        \includegraphics[width=0.241\textwidth]{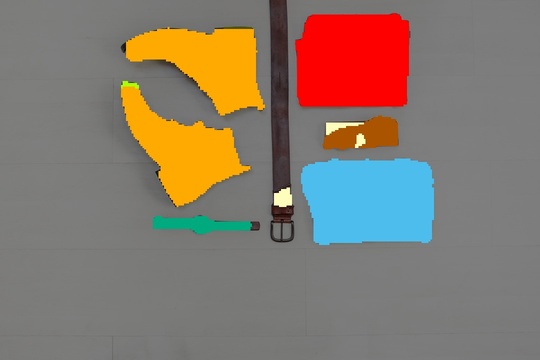}
    \end{subfigure}

    \begin{subfigure}{\textwidth}
        \includegraphics[width=0.24\textwidth]{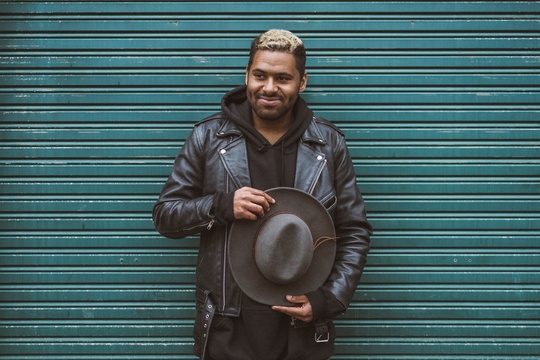}
        \hfill
        \includegraphics[width=0.24\textwidth]{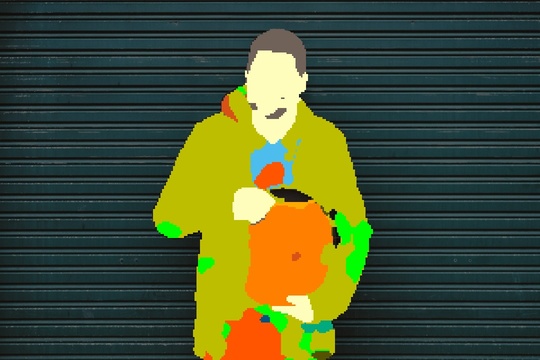}
        \hfill
        \includegraphics[width=0.24\textwidth]{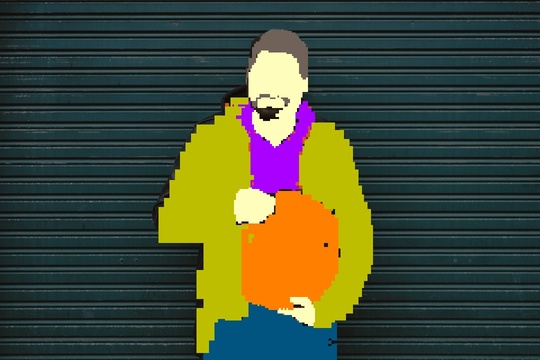}
        \hfill
        \includegraphics[width=0.24\textwidth]{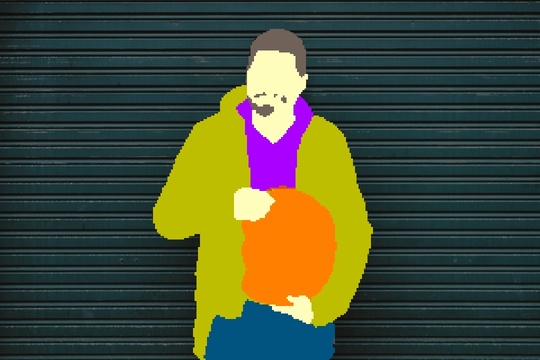}
    \end{subfigure}
    
    \begin{subfigure}{\textwidth}
        \includegraphics[width=0.24\textwidth]{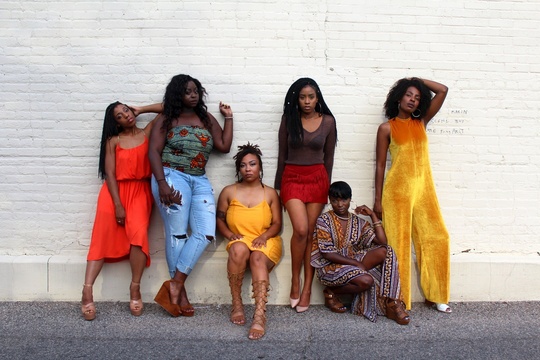}
        \hfill
        \includegraphics[width=0.24\textwidth]{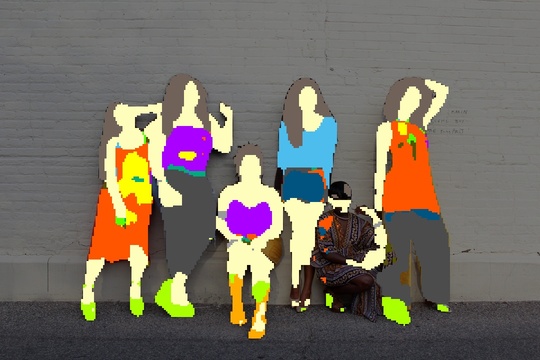}
        \hfill
        \includegraphics[width=0.24\textwidth]{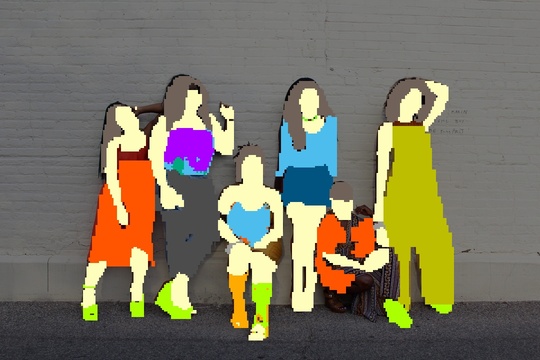}
        \hfill
        \includegraphics[width=0.24\textwidth]{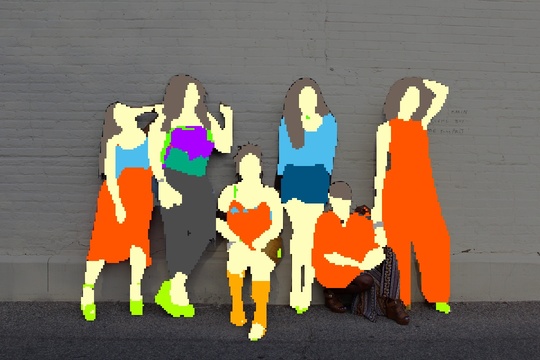}
    \end{subfigure}
    
   \begin{subfigure}{0.49\textwidth}
        \centering
        \begin{subfigure}[b]{0.241\textwidth}
             \includegraphics[width=\textwidth]{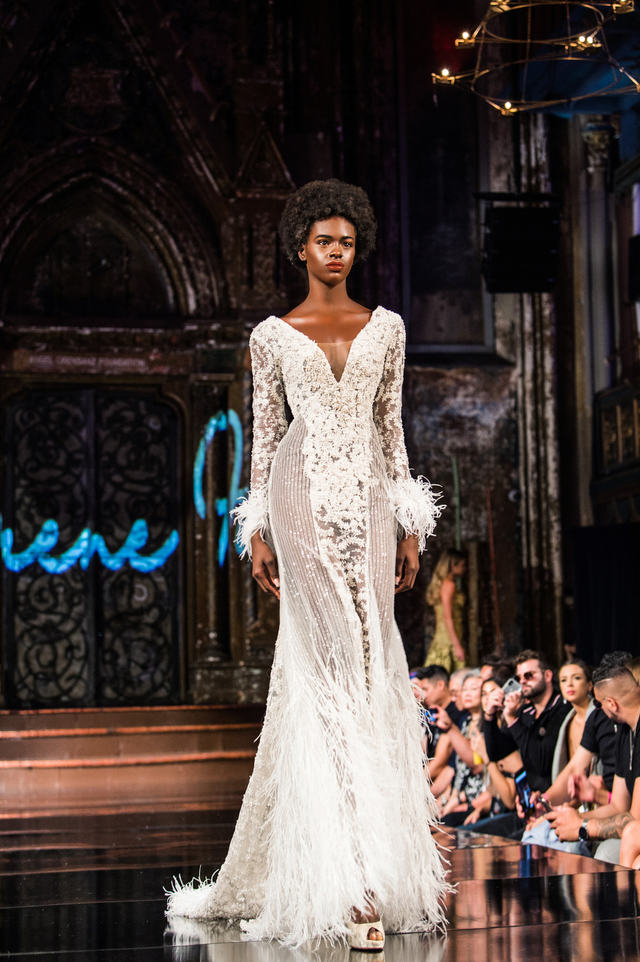}
        \end{subfigure}
        \hfill
        \begin{subfigure}[b]{0.241\textwidth}
            \includegraphics[width=\textwidth]{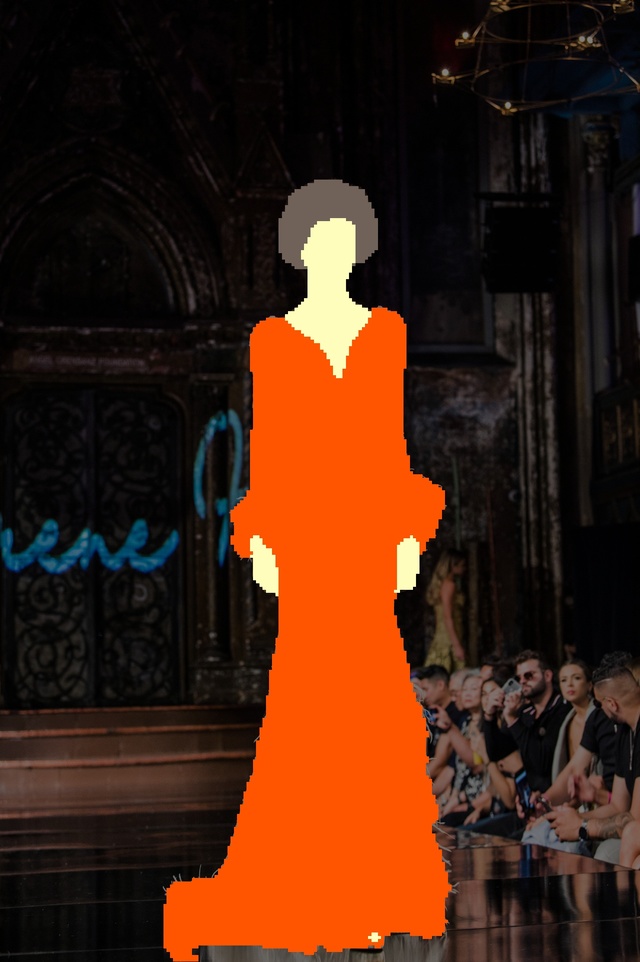}
        \end{subfigure}
        \hfill
        \begin{subfigure}[b]{0.241\textwidth}
            \includegraphics[width=\textwidth]{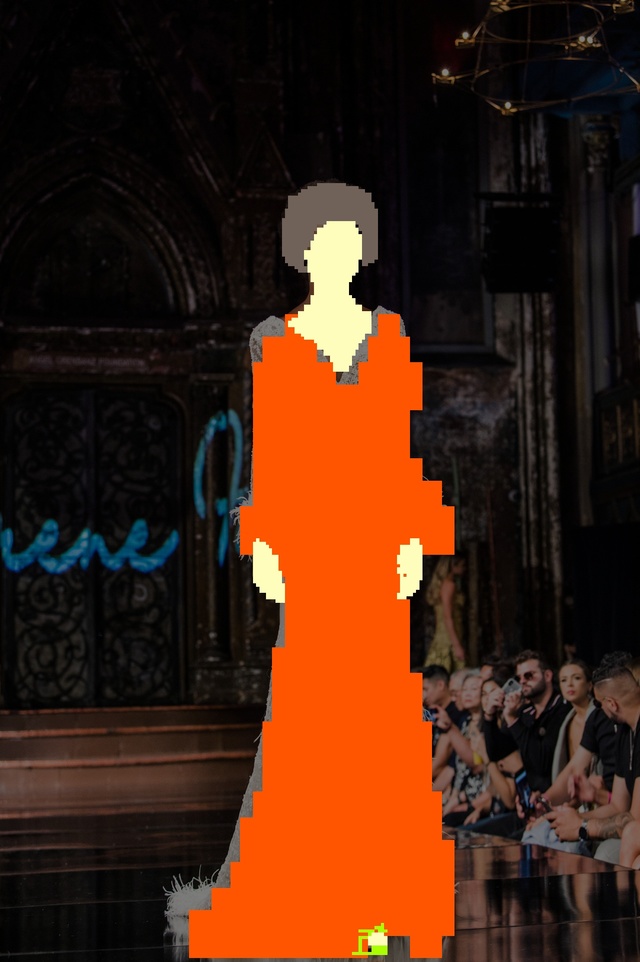}
        \end{subfigure}
        \hfill
        \begin{subfigure}[b]{0.241\textwidth}
            \includegraphics[width=\textwidth]{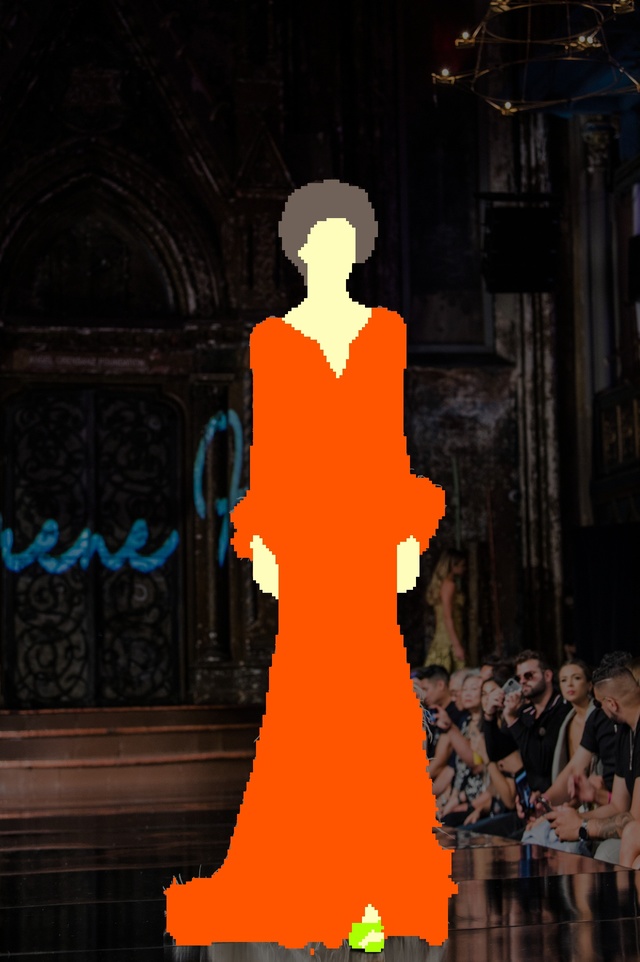}
        \end{subfigure}    
    \end{subfigure}
    \hfill
    \begin{subfigure}{0.49\textwidth}
        \centering
        \begin{subfigure}[b]{0.241\textwidth}
             \includegraphics[width=\textwidth]{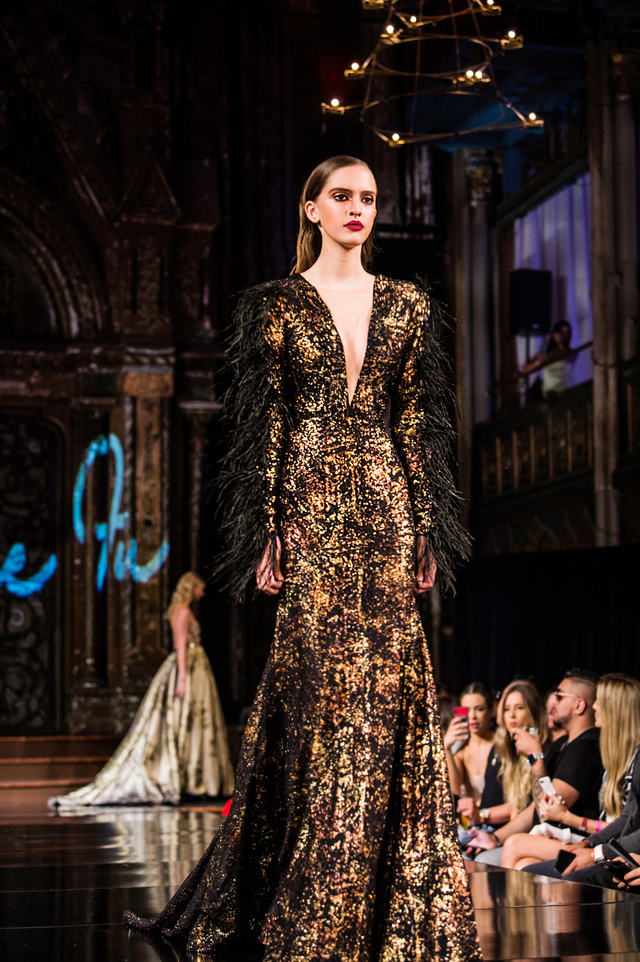}
        \end{subfigure}
        \hfill
        \begin{subfigure}[b]{0.241\textwidth}
            \includegraphics[width=\textwidth]{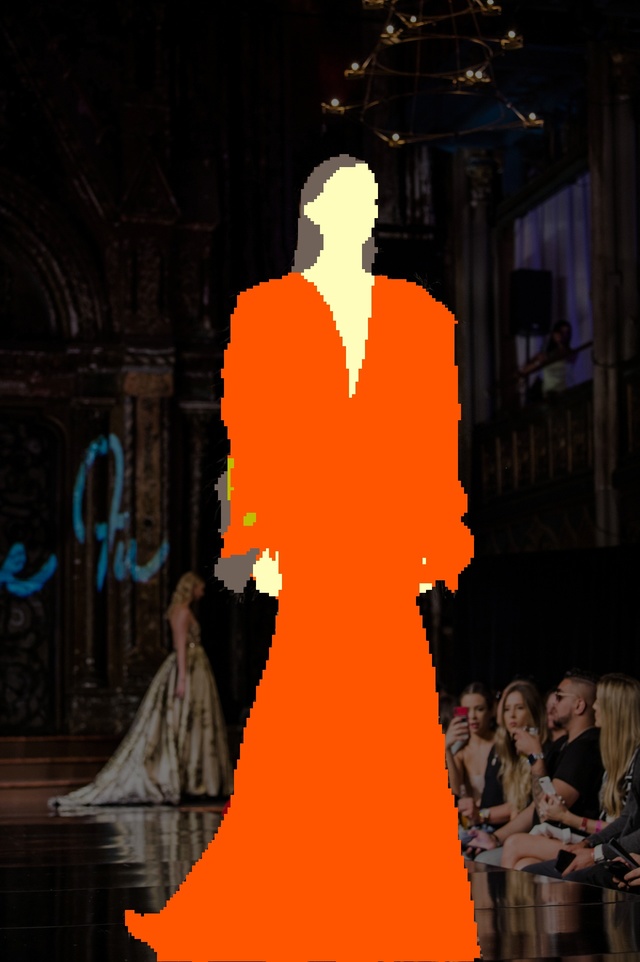}
        \end{subfigure}
        \hfill
        \begin{subfigure}[b]{0.241\textwidth}
            \includegraphics[width=\textwidth]{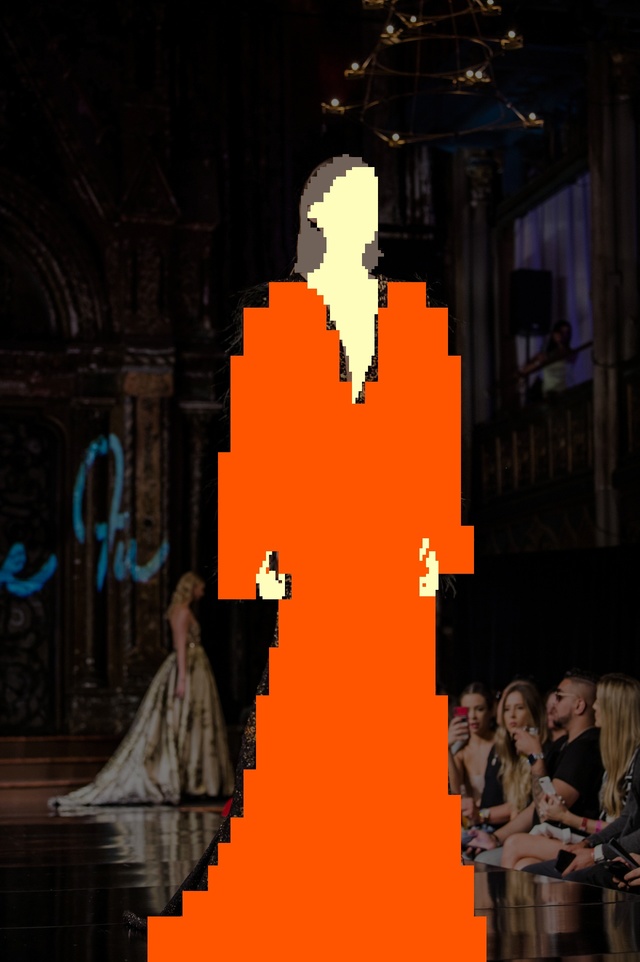}
        \end{subfigure}
        \hfill
        \begin{subfigure}[b]{0.241\textwidth}
            \includegraphics[width=\textwidth]{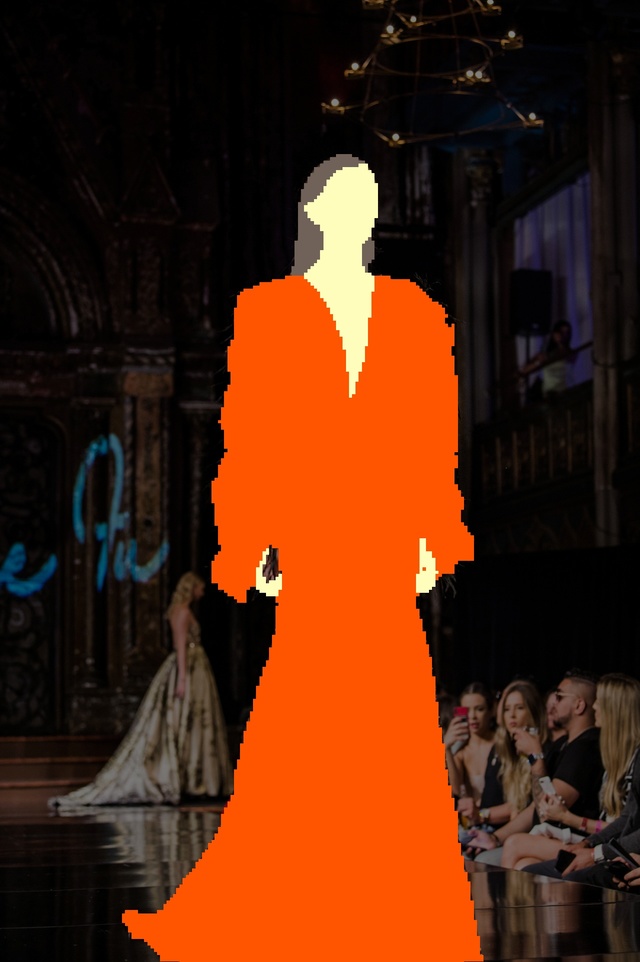}
        \end{subfigure}    
    \end{subfigure}

    \begin{subfigure}{0.49\textwidth}
        \includegraphics[width=0.24\textwidth]{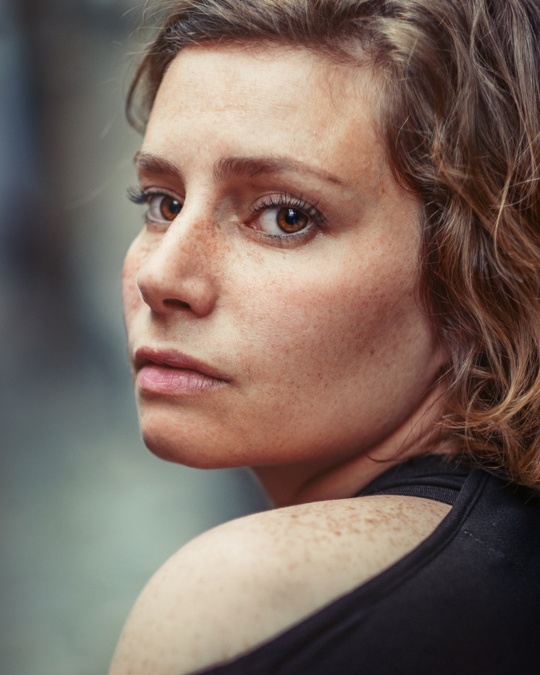}
        \hfill
        \includegraphics[width=0.24\textwidth]{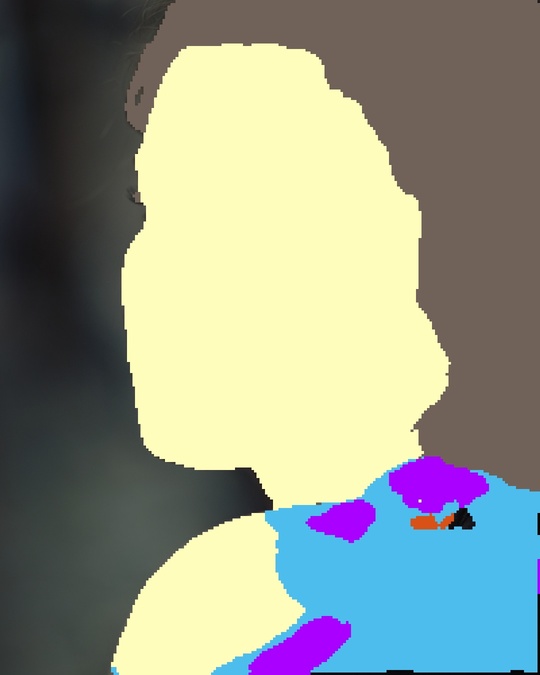}
        \hfill
        \includegraphics[width=0.24\textwidth]{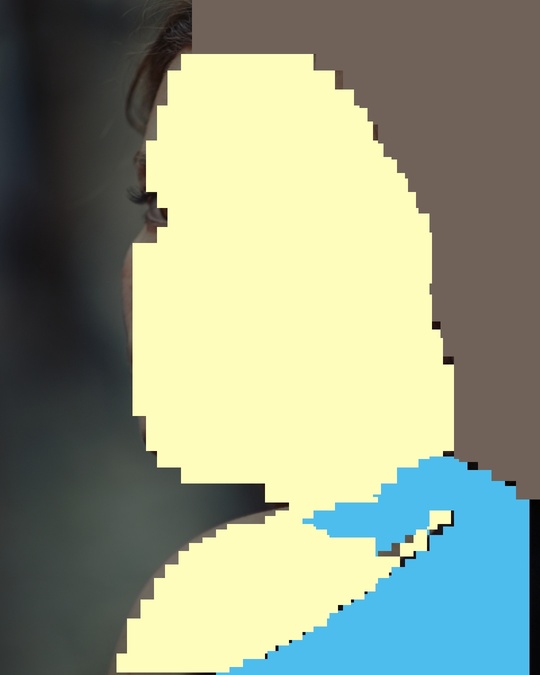}
        \hfill
        \includegraphics[width=0.24\textwidth]{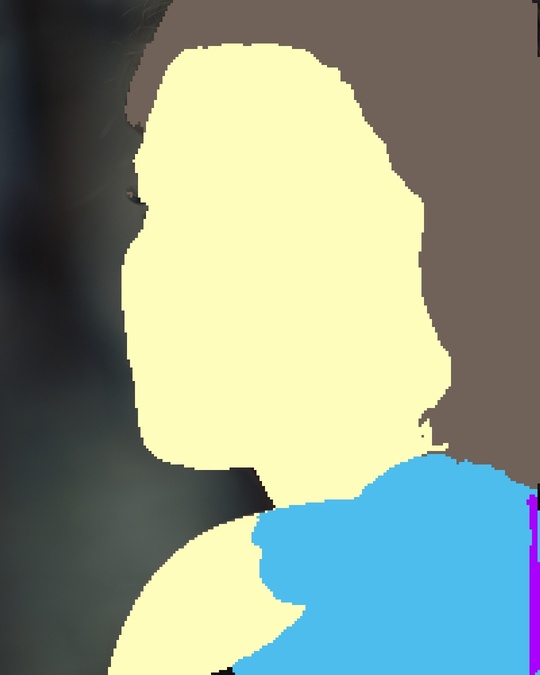}
    \end{subfigure}
    \hfill
    \begin{subfigure}{0.49\textwidth}
        \includegraphics[width=0.24\textwidth]{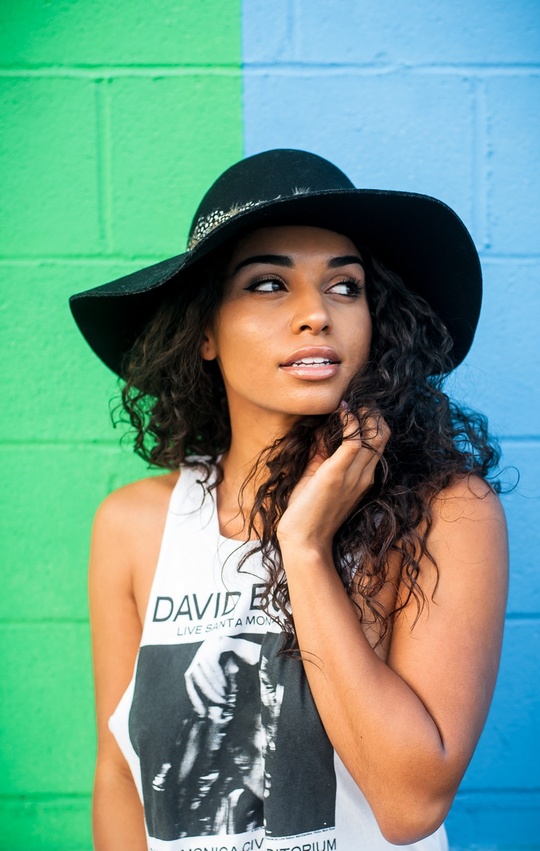}
        \includegraphics[width=0.24\textwidth]{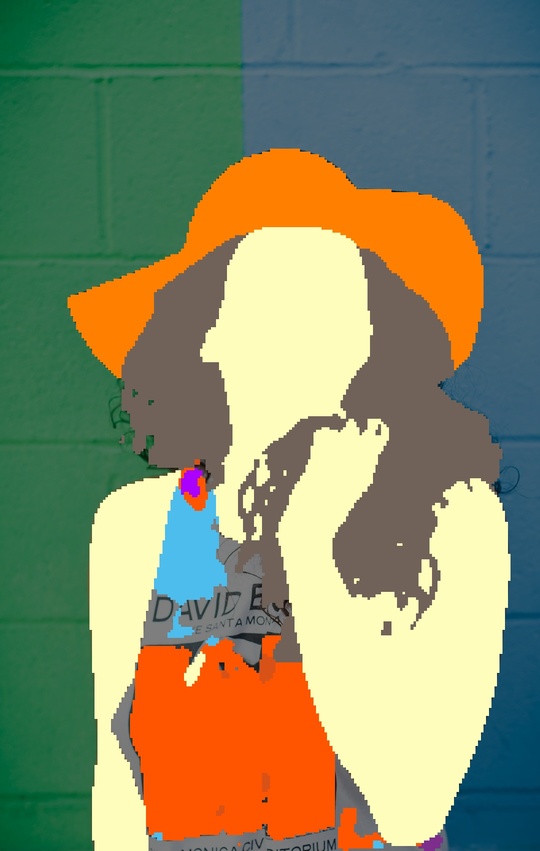}
        \includegraphics[width=0.24\textwidth]{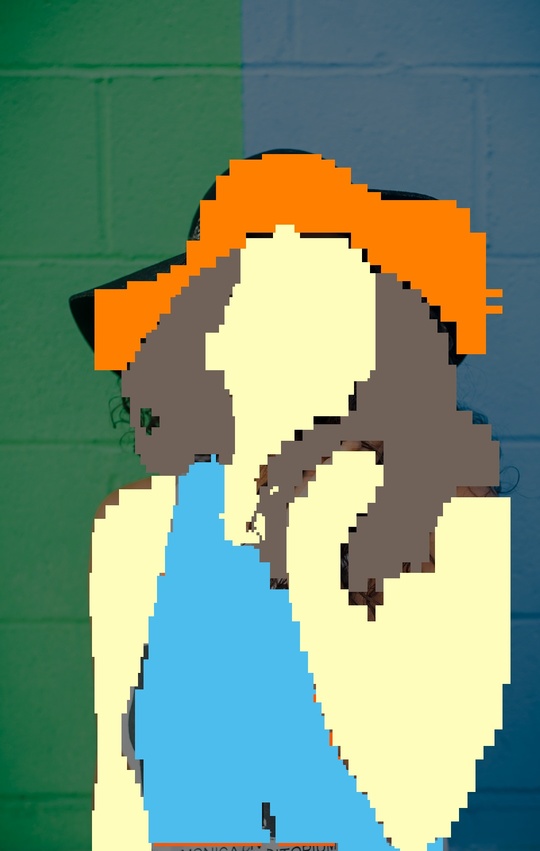}
        \includegraphics[width=0.24\textwidth]{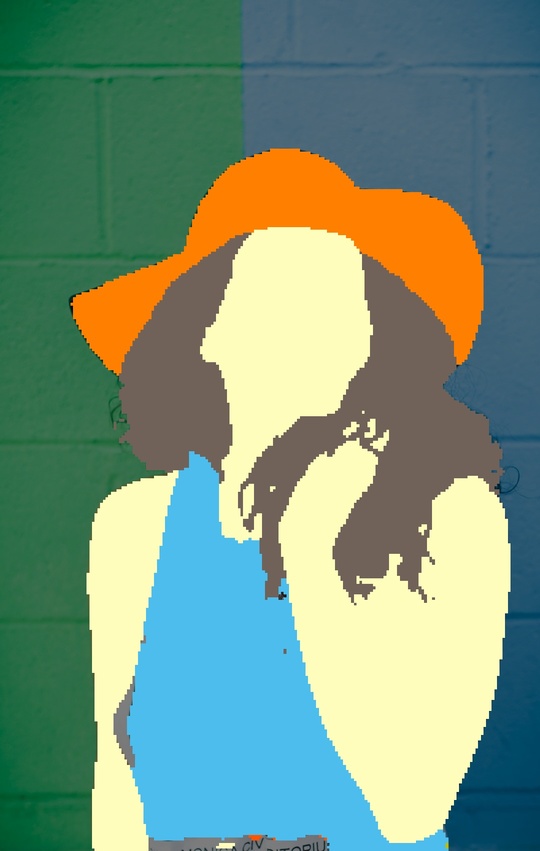}
    \end{subfigure}


    \begin{tikzpicture}[y=.5cm, x=.5cm]
	legend


    \draw[ fill={rgb,255:red,255; green,254; blue,189} ] (0,0) rectangle (1, 1);
    \draw (1, 0.5) node[right, font=\small]{skin};
    
    \draw[ fill={rgb,255:red,112; green,98; blue,90} ] (3,0) rectangle (4, 1);
    \draw (4, 0.5) node[right,font=\small]{hair};
    
    \draw[ fill={rgb,255:red,255; green,128; blue,0} ] (6,0) rectangle (7, 1);
    \draw (7, 0.5) node[right,font=\small]{hat};
    
    \draw[ fill={rgb,255:red,217; green,83; blue,25} ] (9,0) rectangle (10, 1);
    \draw (10, 0.5) node[right,font=\small]{tie};    
    
    \draw[ fill={rgb,255:red,170; green,85; blue,0} ] (12,0) rectangle (13, 1);
    \draw (13, 0.5) node[right,font=\footnotesize]{glasses};
    
    \draw[ fill={rgb,255:red,84; green,255; blue,0}  ] (15,0) rectangle (16, 1);
    \draw (16, 0.5) node[right,font=\footnotesize]{necklace};

    \draw[ fill={rgb,255:red,170; green,255; blue,0} ] (19, 0) rectangle (20, 1);
    \draw (20, 0.5) node[right,font=\small]{shoes};
    
    \draw[ fill={rgb,255:red,255; green,170; blue,0} ] (22, 0) rectangle (23, 1);
	\draw (23, 0.5) node[right,font=\small]{boots};
	
	\draw[ fill={rgb,255:red,77; green,77; blue,77}  ] (25,0) rectangle (26, 1);
	\draw (26, 0.5) node[right,font=\small]{pants};
	
	\draw[ fill={rgb,255:red,0; green,0; blue,255} ] (28, 0) rectangle (29, 1);
    \draw (29, 0.5) node[right,font=\footnotesize]{leggings};
    
    \draw[ fill={rgb,255:red,255; green,255; blue,0} ] (31, 0) rectangle (32, 1);
    \draw (32, 0.5) node[right,font=\footnotesize]{jumpsuit};

    \draw[ fill={rgb,255:red,77; green,190; blue,238} ] (0,-1) rectangle (1, 0);
    \draw (1, -0.5) node[right,font=\small]{t-shirt};
    
    \draw[ fill={rgb,255:red,170; green,0; blue,255} ] (3,-1) rectangle (4, 0);
    \draw (4, -0.5) node[right,font=\small]{shirt};
    
    \draw[ fill={rgb,255:red,255; green,84; blue,0} ] (6,-1) rectangle (7, 0);
    \draw (7, -0.5) node[right,font=\small]{dress};
    
    \draw[ fill={rgb,255:red,191; green,191; blue,0} ] (9,-1) rectangle (10, 0);
    \draw (10, -0.5) node[right,font=\small]{jacket};
    
    \draw[ fill={rgb,255:red,0; green,255; blue,0} ] (12,-1) rectangle (13, 0);
    \draw (13, -0.5) node[right,font=\small]{coat};
    
\end{tikzpicture}
    \caption{ This Figure is an extension of Figure~\ref{fig:vis_cases}. From left to right, images, results of Panoptic-FPN, results of Mask R-CNN-IMP, results of our final model, Panoptic-FPN-IMP. The proposed method, IMP, works well on different types of clothing parsing examples, from vintage images, layflat images, street-fashion examples, fashion-runway photos, and photos with full or partial-bodies visible.  
    }
    \label{fig:vis_cases_sup}
    \vspace{-3.0mm}
\end{figure*}

In Figure~\ref{fig:vis_cases_sup}, we show more  qualitative examples besides Figure~\ref{fig:vis_cases}. We use ResNet-50-FPN as the backbone model and train the model on the ~\Shopagon. Figure~\ref{fig:vis_cases_sup} contains more diverse photos, such as vintage photos, layflat photos and images with full or half-bodies visible. Although Mask R-CNN-IMP can generate cleaner results than Panoptic-FPN, Mask R-CNN-IMP also incurs poor performance on boundaries of large objects which was caused by the low resolution output of Mask R-CNN\footnote{28$\times$28}. Our final model Panoptic-FPN-IMP can generate sharp semantic segmentation results but also makes labeling of pixels from the same objects consistent. 

\begin{table}[h]
    \begin{tabular}{c|c|c|r|r}
         class & \multicolumn{2}{c}{Difference} & \#Instances & Total area \\
               & & DA & \\ 
         \hline 
         Person & 0.7 &  1.1  & 17,395     & 64,901,113\\
         Rider & 4.4  &  5.1      & 1,660  & 7,169,330 \\
         Car & 0.2    &  0.4      & 26,180 & 380,112,819 \\
         Truck & 2.5  &  9.4     & 466    & 14,657,648\\
         Bus & 1.2    &  6.9       & 350    & 12,684,337 \\
         Train & 9.8  &  5.5     & 158    & 11,643,940\\ 
         Motorcycle & 6.5 & 4.3  & 705    & 5,037,718\\ 
         Bicycle & 0.6 & 0.8   & 3,433  & 14,646,908 \\
         \hline 
         Average & 3.2 & 4.2 
         
    \end{tabular}
    \caption{Analysis of Semantic Segmentation classes which are also Instance Segmentation. There is a correlation if the class has fewer instances and area, it gets more improvement from Instance Mask Projection. \textbf{DA}: with Data Augmentation.}
    \label{tab:cityscape_instance_analysis}
    \vspace{-2.0mm}
\end{table}

\subsection*{More discussions on Cityscapes dataset.}
Table~\ref{tab:cityscape_instance_analysis} shows the mIOU difference of Thing classes of Cityscapes with and without the data augmentation. This Table is part of Table~\ref{tab:cityscape_perclass_iou} but adds number of instances and area information. We found out the improvement is also similar to the clothing datasets. First, the classes with less examples are improved more. See Train(\#158), Bus(\#350), Truck(\#466), and Motorcycle(\#705). Another is the improvement among the confusing classes. Although Rider contains enough examples, its similarity to Person, makes its mIOU lower. Our model is useful to distinguish these cases and increases the mIOU of Rider significantly. 

\begin{figure*}[!t]
   \begin{subfigure}{\textwidth}
        \centering
        \begin{subfigure}[b]{0.24\textwidth}
            \caption*{Image}
            \includegraphics[width=\textwidth]{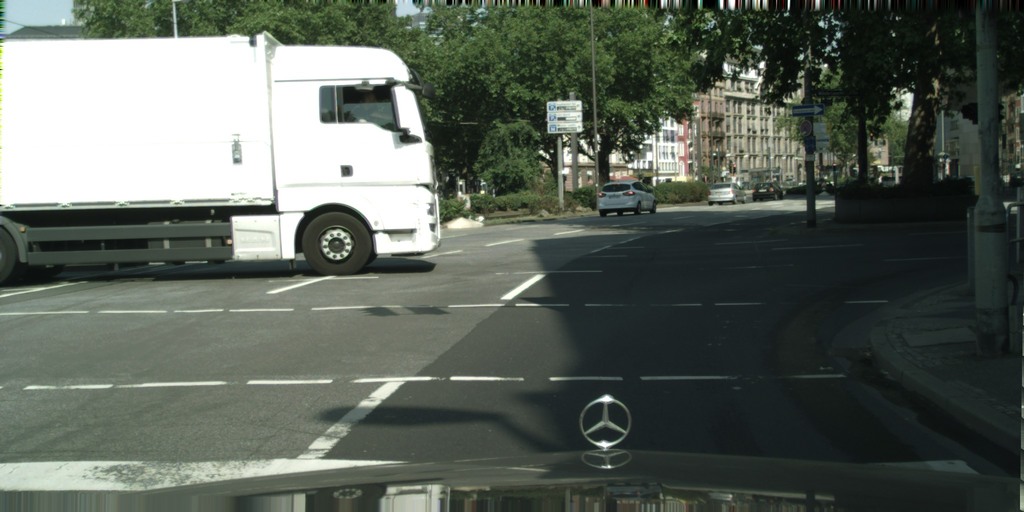}
        \end{subfigure}
        \hfill
        \begin{subfigure}[b]{0.24\textwidth}
            \caption*{Panoptic-FPN}
            \includegraphics[width=\textwidth]{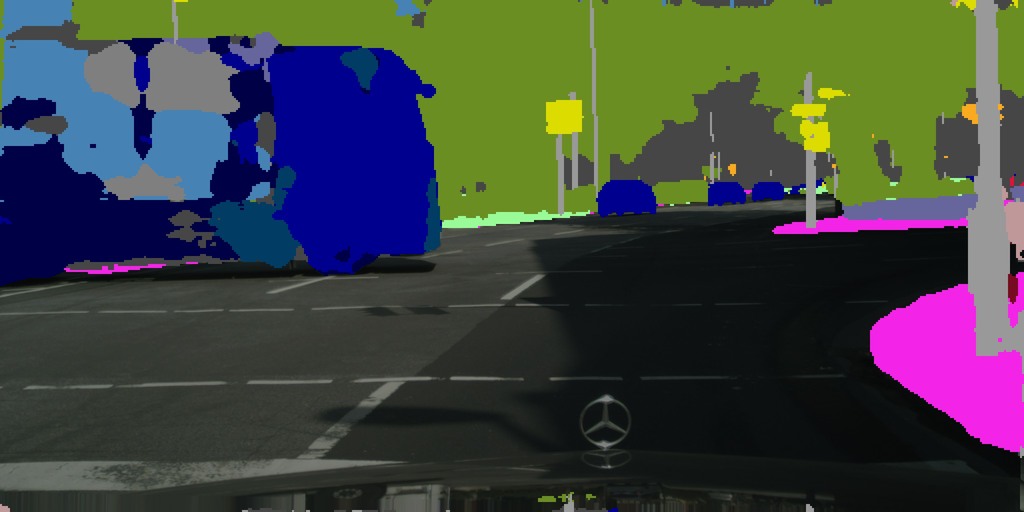}
        \end{subfigure}
        \hfill
        \begin{subfigure}[b]{0.24\textwidth}
            \caption*{Panoptic-FPN-IMP}
            \includegraphics[width=\textwidth]{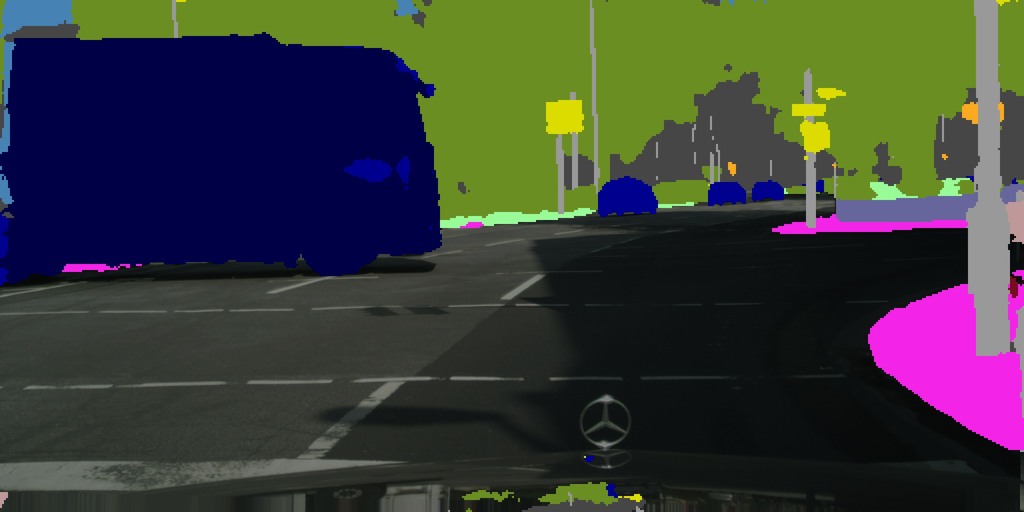}
        \end{subfigure}
        \hfill
        \begin{subfigure}[b]{0.24\textwidth}
            \caption*{GroundTruth}
            \includegraphics[width=\textwidth]{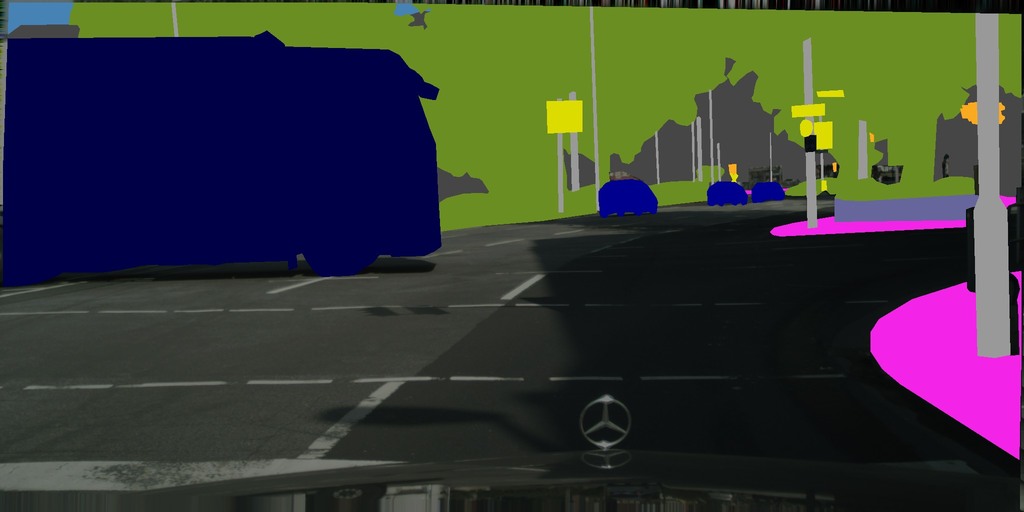}
        \end{subfigure}
    
    \caption{Truck} 
    \label{fig:vis_city_truck}
    \end{subfigure}
    \begin{subfigure}{\textwidth}
        \centering
        \begin{subfigure}[b]{0.245\textwidth}
            \includegraphics[width=\textwidth]{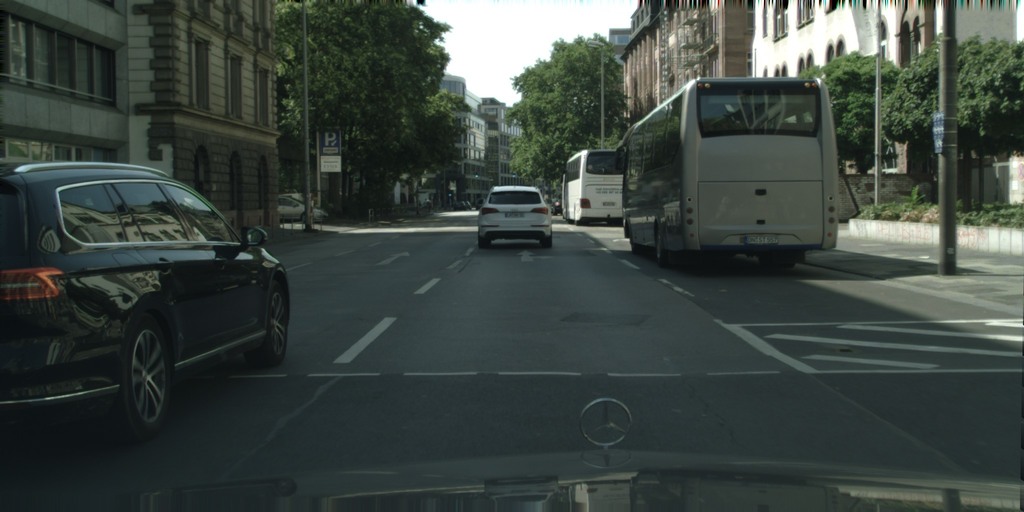}
        \end{subfigure}
        \hfill
        \begin{subfigure}[b]{0.245\textwidth}
            \includegraphics[width=\textwidth]{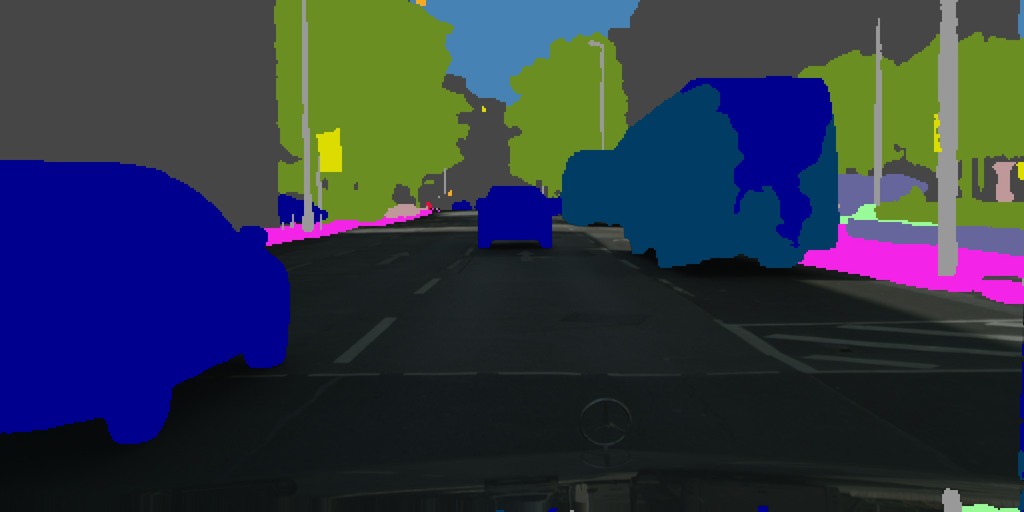}
        \end{subfigure}
        \hfill
        \begin{subfigure}[b]{0.245\textwidth}
            \includegraphics[width=\textwidth]{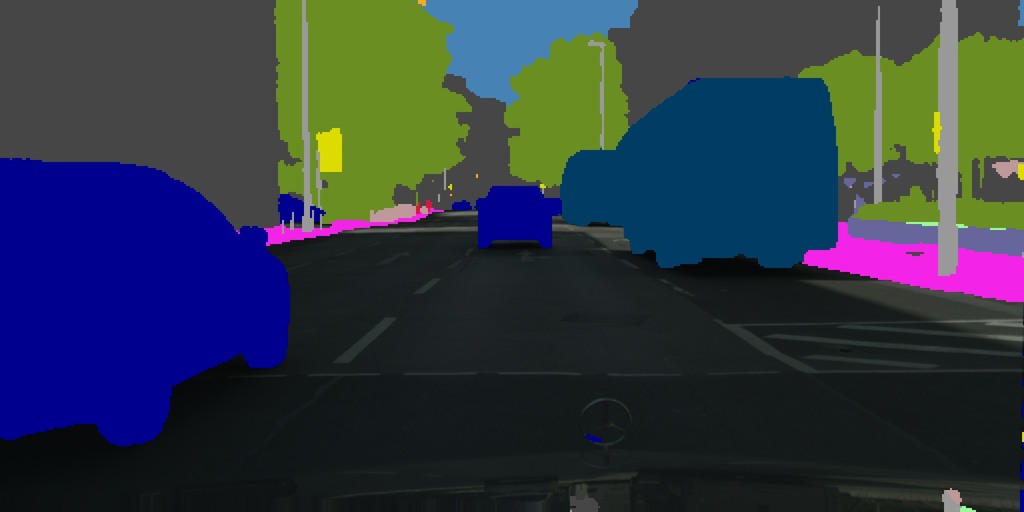}
        \end{subfigure}
        \hfill
        \begin{subfigure}[b]{0.245\textwidth}
            \includegraphics[width=\textwidth]{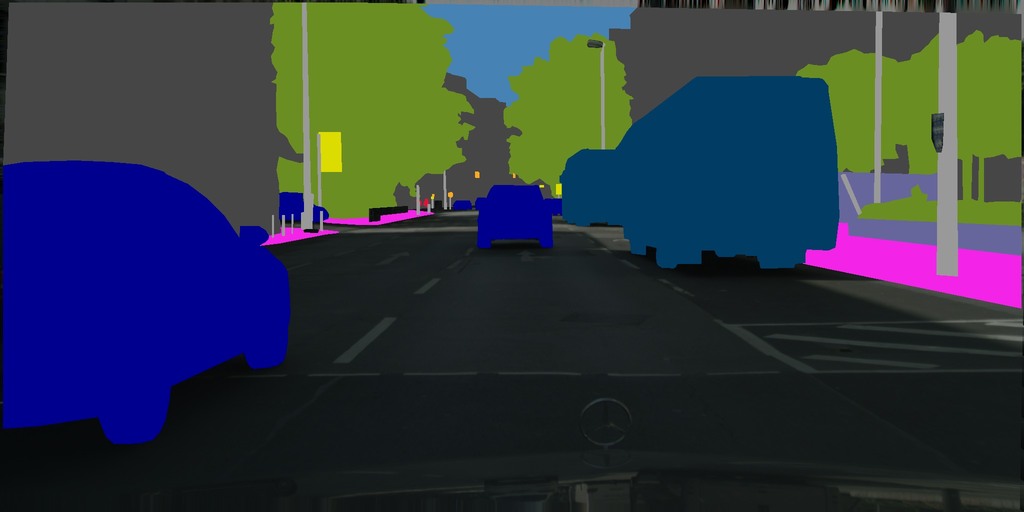}
        \end{subfigure}
        \\
    \caption{Bus} 
    \label{fig:vis_city_bus}
    \end{subfigure}
    \\
    \begin{subfigure}{\textwidth}
        \centering
        \begin{subfigure}[b]{0.245\textwidth}
            \includegraphics[width=\textwidth]{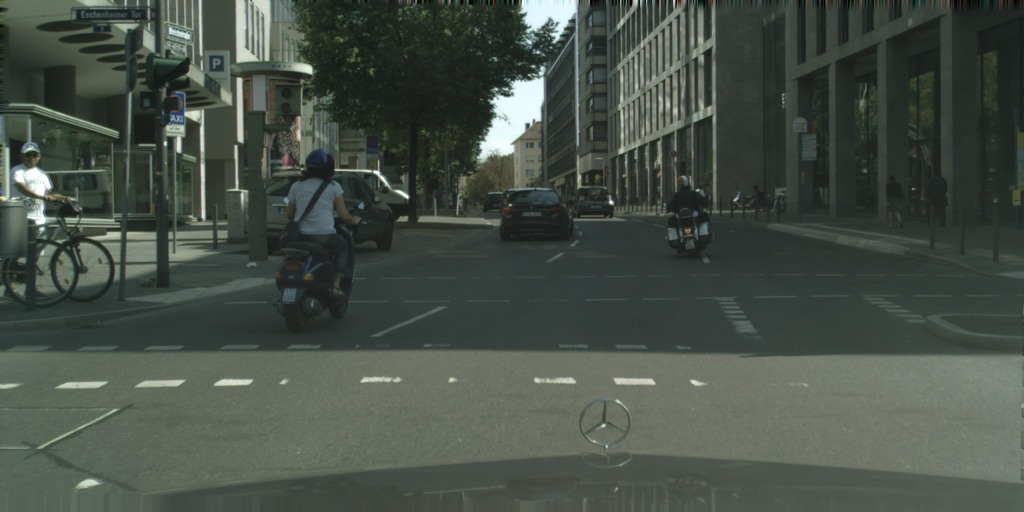}
        \end{subfigure}
        \hfill
        \begin{subfigure}[b]{0.245\textwidth}
            \includegraphics[width=\textwidth]{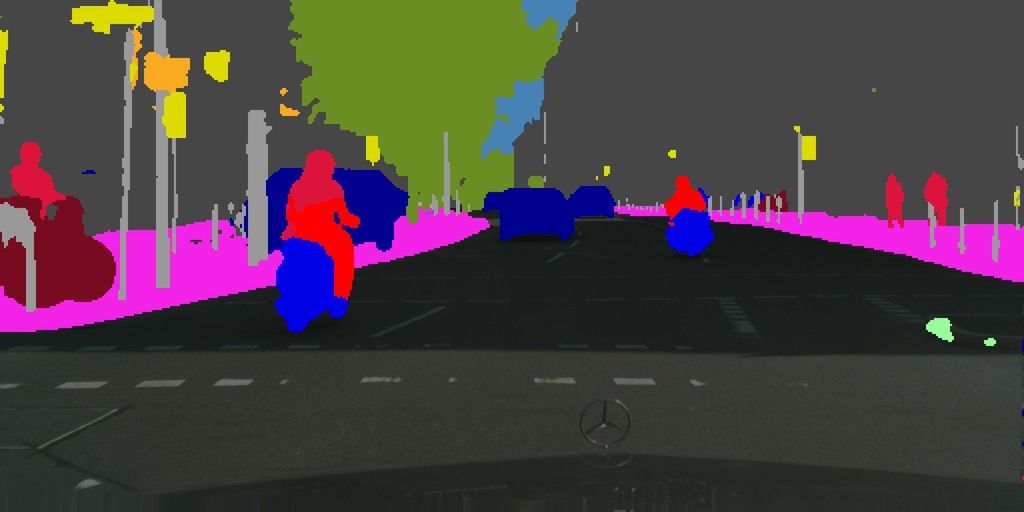}
        \end{subfigure}
        \hfill
        \begin{subfigure}[b]{0.245\textwidth}
            \includegraphics[width=\textwidth]{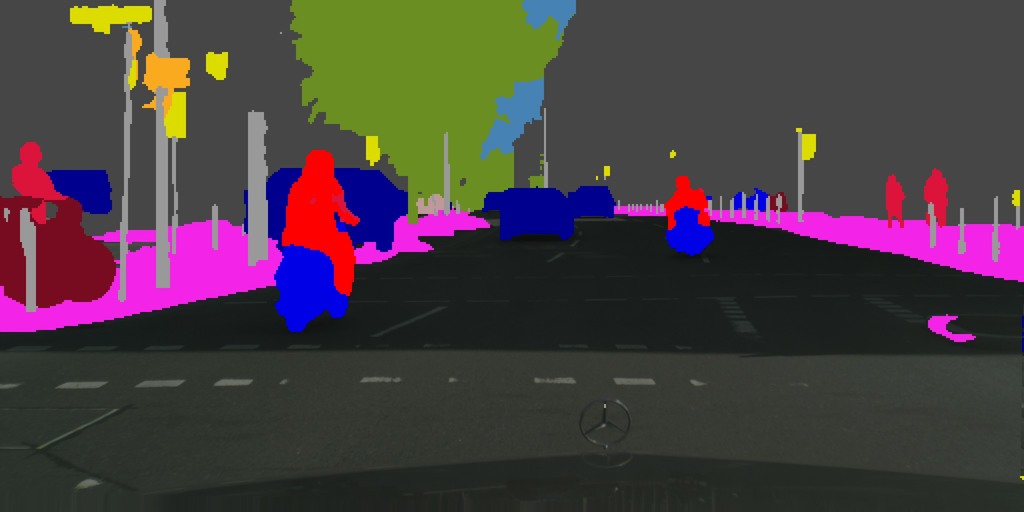}
        \end{subfigure}
        \hfill
        \begin{subfigure}[b]{0.245\textwidth}
            \includegraphics[width=\textwidth]{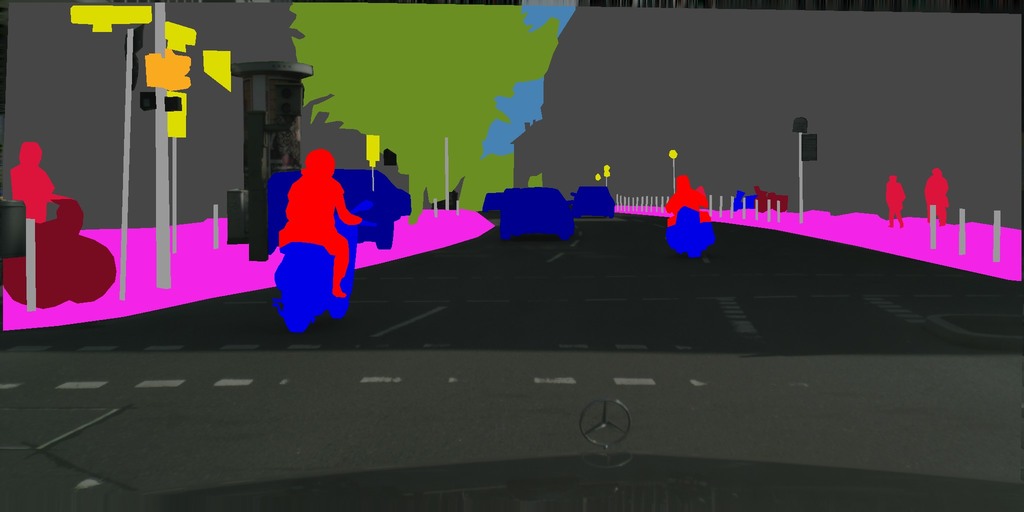}
        \end{subfigure}
        \\
    \caption{Rider} 
    \label{fig:vis_city_rider1}
    \end{subfigure}
    \\
    \begin{subfigure}{\textwidth}
        \centering
        \begin{subfigure}[b]{0.245\textwidth}
            \includegraphics[width=\textwidth]{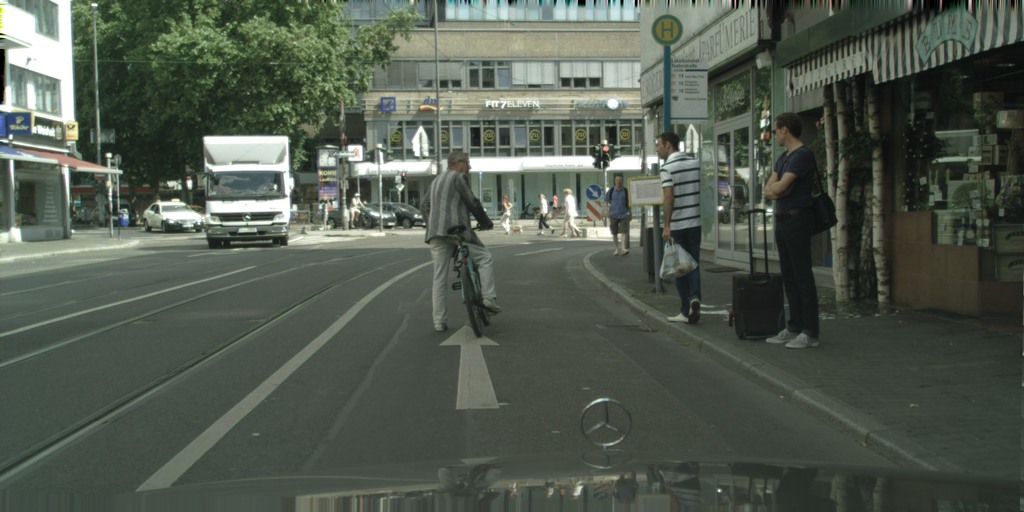}
        \end{subfigure}
        \hfill
        \begin{subfigure}[b]{0.245\textwidth}
            \includegraphics[width=\textwidth]{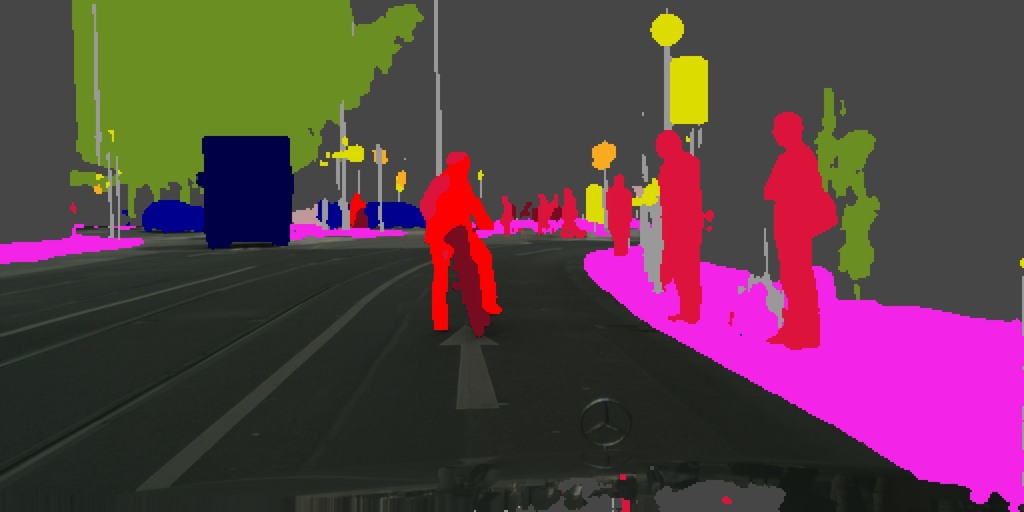}
        \end{subfigure}
        \hfill
        \begin{subfigure}[b]{0.245\textwidth}
            \includegraphics[width=\textwidth]{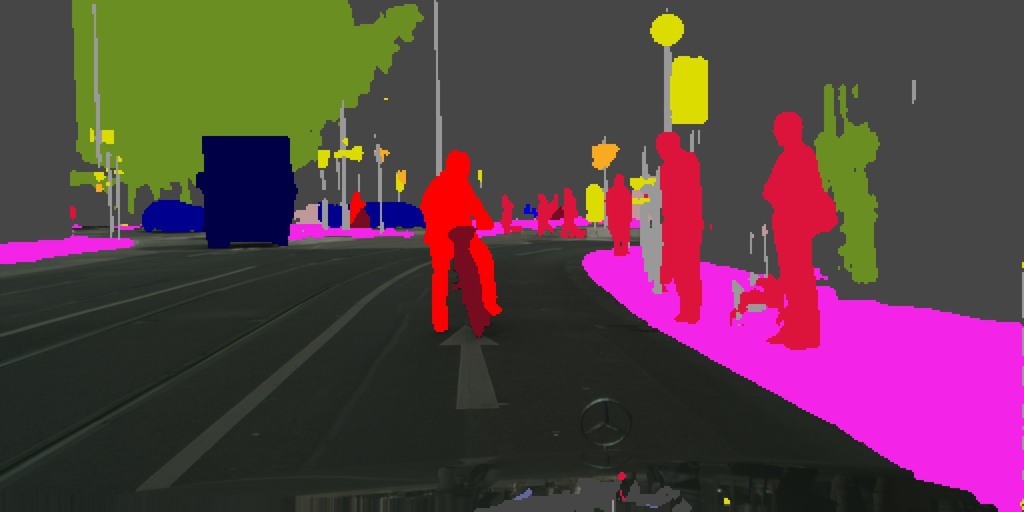}
        \end{subfigure}
        \hfill
        \begin{subfigure}[b]{0.245\textwidth}
            \includegraphics[width=\textwidth]{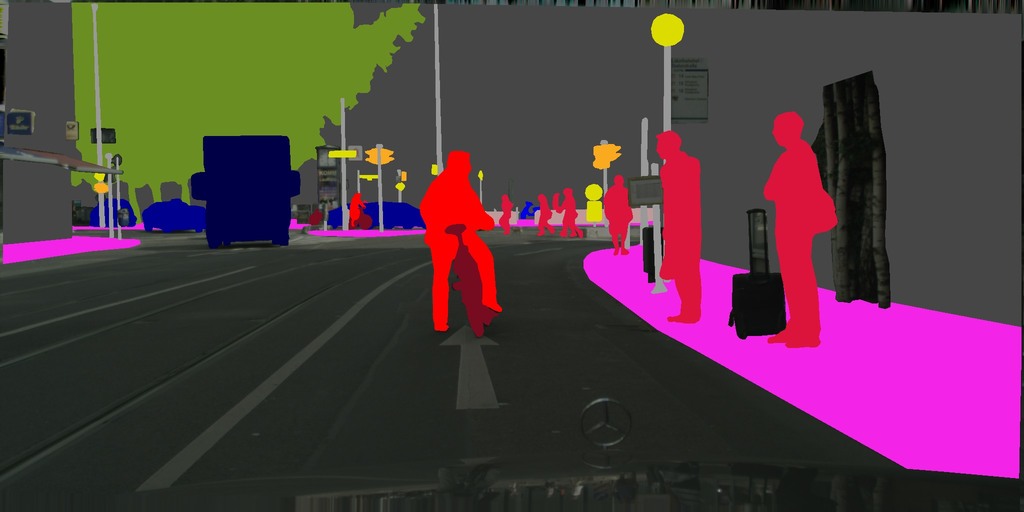}
        \end{subfigure}
        \\
    \caption{Rider} 
    \label{fig:vis_city_rider2}
    \end{subfigure}
    \\
    
    \caption{From left to right, images, results of Panoptic-FPN, Panoptic-FPN-IMP and GroundTruth. With the Instance Mask Projection, our final model, shows cleaner results on Truck(a), Bus(b), and Rider(c,d) classes. 
    }
    \label{fig:vis_city}
    \vspace{-3.0mm}
\end{figure*}

Figure~\ref{fig:vis_city} shows the visualization examples of results of our models. We found that the qualitative results are also similar to the clothing datasets. Our final model, Panoptic-FPN-IMP, provides leaner results. See the better results of segments of Bus and Truck in  Figure~\ref{fig:vis_city_truck} and ~\ref{fig:vis_city_bus}. Another interesting case is Rider which means the person on the motorcycle or bicycle. The top part of Rider of Panoptic-FPN  in Figure ~\ref{fig:vis_city_rider1} and ~\ref{fig:vis_city_rider2} are misclassified as Person. But with Instance Mask Projection, our final model shows correct labeling of all pixels of Rider.

\clearpage

\clearpage


{\small
\bibliographystyle{ieee}
\bibliography{egbib}
}

\end{document}